\documentclass[11pt]{article}

\usepackage[final]{acl}

\usepackage{times}
\usepackage{latexsym}
\usepackage[utf8]{inputenc}
\usepackage{csquotes}
\usepackage{graphicx}

\usepackage[T1]{fontenc}

\usepackage[utf8]{inputenc}

\usepackage{microtype}

\usepackage{inconsolata}

\usepackage{graphicx}
\usepackage{booktabs}
\usepackage{multicol}
\usepackage{multirow}
\usepackage{amsmath}
\usepackage{amsfonts}

\title{Alignment Data Map for \\
Efficient Preference Data Selection and Diagnosis}

\makeatletter
\def\@fnsymbol#1{%
  \ifcase#1\or
    $\dagger$\or
    *\or
    \ddagger\or
    \mathsection\or
    \mathparagraph\or
    \|\or
    **\or
    \dagger\dagger\or
    \ddagger\ddagger
  \else
    \@ctrerr
  \fi
}
\makeatother

\author{%
  Seohyeong Lee$^{1,}$\thanks{\; Equal contribution}\;\;
  Eunwon Kim$^{2,}$\footnotemark[1]\;\;
  Hwaran Lee$^{1,}$\thanks{\; Corresponding author}\;\;
  Buru Chang$^{3,}$\footnotemark[2]\\
  \textsuperscript{1}Sogang University\;\;
  \textsuperscript{2}Upstage\;\;
  \textsuperscript{3}Korea University\\
  \texttt{\{seohyeong, hwaranlee\}@sogang.ac.kr} \;\;
  \texttt{eunwon@upstage.ai} \;\;\\
  \texttt{buru\_chang@korea.ac.kr}
}

\begin{document}
\maketitle
\begin{abstract}\label{sec:0_abstract}
Human preference data is essential for aligning large language models (LLMs) with human values, but collecting such data is often costly and inefficient---motivating the need for efficient data selection methods that reduce annotation costs while preserving alignment effectiveness.
To address this issue, we propose \textit{Alignment Data Map}, a data analysis tool for identifying and selecting effective preference data.
We first evaluate alignment scores of the preference data by LLM-as-a-judge, explicit reward model, and reference-based approaches.
The Alignment Data Map considers both response quality and inter-response variability based on the alignment scores.
From our experimental findings, training on only 33\% of samples that exhibit high-quality and low-variability achieves comparable or superior alignment performance on MT-Bench, Evol-Instruct, and AlpacaEval, compared to training with the full dataset.
In addition, Alignment Data Map detects potential label misannotations by analyzing correlations between annotated labels and alignment scores, improving annotation accuracy.
The implementation is available at \url{https://github.com/01choco/Alignment-Data-Map}
\end{abstract}
\section{Introduction}\label{sec:1_introduction}
In LLM alignment learning~\cite{ouyang2022training}, human preference datasets serve as a key resource, typically consisting of response candidates with preference or ranking feedback given to a user instruction. 
However, constructing high-quality preference datasets faces scalability challenges due to the prohibitive cost and complexity of human annotation~\cite{kopf2024openassistant, sun2024principle}. 
Given these constraints, identifying which samples contribute most significantly to alignment has become essential to improve the efficiency of both data collection and alignment learning.

\begin{figure*}[!ht]
    \centering
    \includegraphics[width=\textwidth]{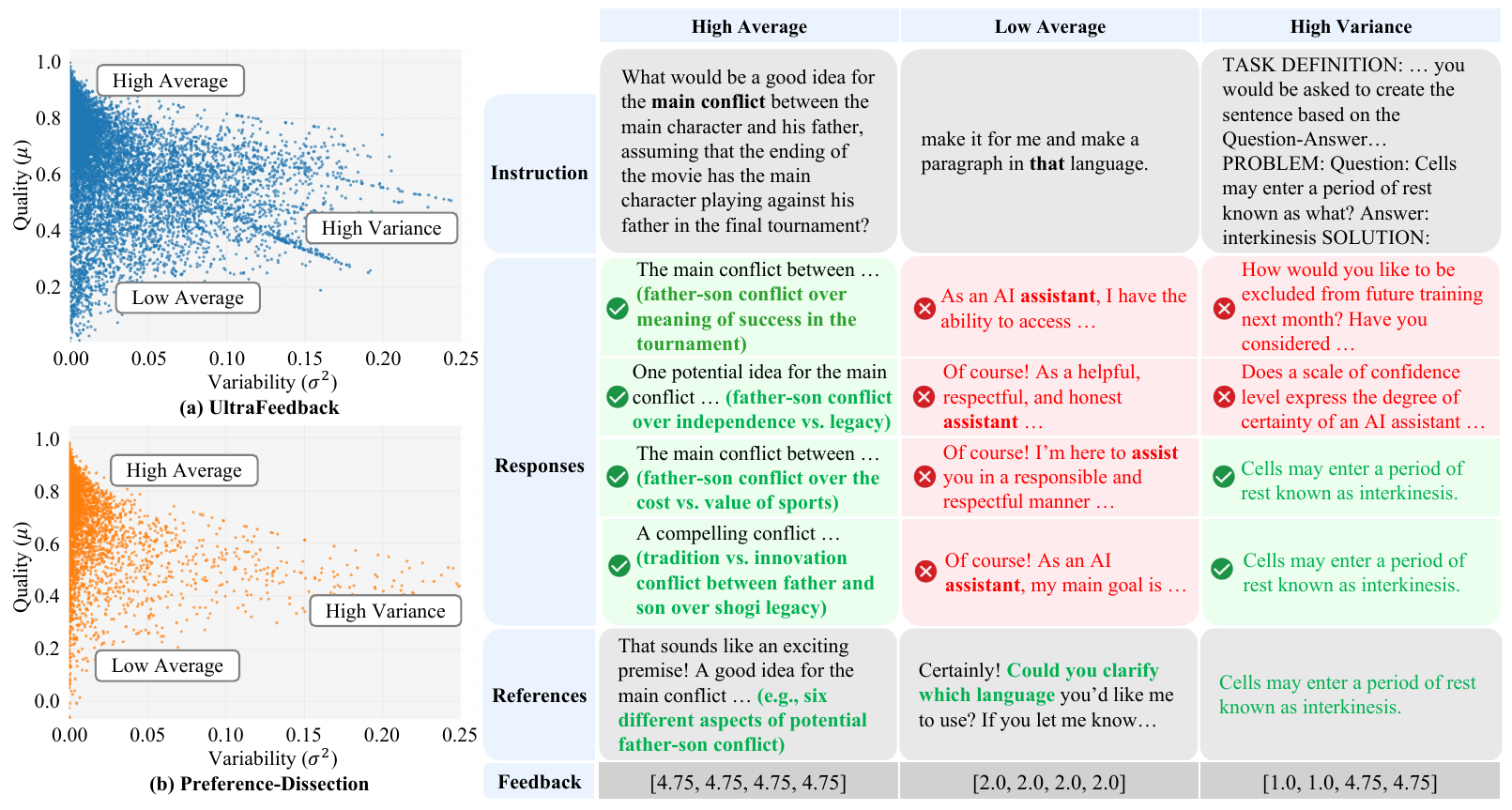}
    \caption{
        (Left) Alignment Data Maps for (a) UltraFeedback~\cite{cui2024ultrafeedback} and (b) Preference-Dissection~\cite{li-etal-2024-dissecting}. (Right) Representative response examples from three regions.
The High Average region contains consistently high-quality responses with diverse writing styles.
The Low Average region includes generic and low-quality responses.
The High Variance region exhibits clear quality disparities across responses, resulting in low ambiguity in preference judgments.
    }
    \label{fig:1_plot_sample}
    \vspace{-1em}
\end{figure*}

Recent studies have largely focused on the \textit{reward margin}, defined as the difference in reward scores between competing response candidates under a given instruction, as a primary signal for identifying effective preference data.
In particular, several preference optimization and data selection approaches prioritize samples with small reward margins, based on the intuition that ambiguous preference comparisons provide stronger learning signals for alignment~\cite{stiennon2020learning,muldrew2024active,yang2024not}.

However, we argue that the reward margin alone provides an incomplete signal for effective data selection; rather, the absolute quality of responses must also be taken into account.
In DPO-style preference alignment, prior work proposes that high-quality (\textit{i.e.}, high reward score) responses are more effective for alignment, as they provide more helpful supervision signals~\cite{pan2025what}.
Yet, reward margin fails to capture this distinction: data samples with identical margins can differ substantially in the absolute quality of their response candidates.
As illustrated in Figure~\ref{fig:1_plot_sample}, low-margin samples may consist of high-quality responses (\textit{i.e., High Average}) or poorly aligned ones (\textit{i.e., Low Average}).
These observations indicate that margin-based selection alone cannot reliably distinguish effective preference data, necessitating an approach that explicitly accounts for response quality.

Motivated by this observation, we introduce \textit{Alignment Data Map}, a data analysis tool that maps and diagnoses preference data from an alignment perspective by evaluating response \textit{variability} and \textit{quality} --- the variance and average of alignment scores, respectively.
Here, we evaluate the alignment score of each response candidate using various methods: an LLM-as-a-judge~\cite{zheng2023judging}, explicit reward model~\cite{ouyang2022training}, and reference-based score~\cite{zhang2019bertscore}.
Moreover, our variability metric generalizes beyond pairwise comparisons to capture margin-based signals across multiple response candidates.
By jointly considering variability and quality, our approach enables more reliable data selection by separating samples with similar variability but different quality levels.
Empirically, we demonstrate that training on only 33\% of data in the High Average region of the Alignment Data Map achieves comparable or even superior alignment performance compared to the full dataset across MT-Bench~\cite{zheng2023judging}, Evol-Instruct~\cite{xu2024wizardlm}, and AlpacaEval~\cite{alpaca_eval}.

Furthermore, we show that Alignment Data Map serves as a diagnostic tool for identifying noise in preference annotations.
By analyzing the correlation between the collected labels and alignment scores, we reveal systematic mismatches that effectively flag unreliable data points.

In summary, our key contributions are:
\begin{itemize}
    \item 
    We introduce \textit{Alignment Data Map}, a data analysis tool that organizes preference data by jointly considering response variability and quality, enabling clearer separation of effective preference learning data.
    \item 
    We present a diagnostic approach for preference data that supports both data selection and the validation of preference labels based on their consistency with alignment scores.
    \item We conduct comprehensive and reproducible experiments demonstrating that Alignment Data Map improves data collection efficiency and facilitates effective quality analysis of labeled preference data.
\end{itemize}

\section{Alignment Dataset Cartography}
\begin{figure}[t]
    \centering
    \includegraphics[width=\columnwidth]{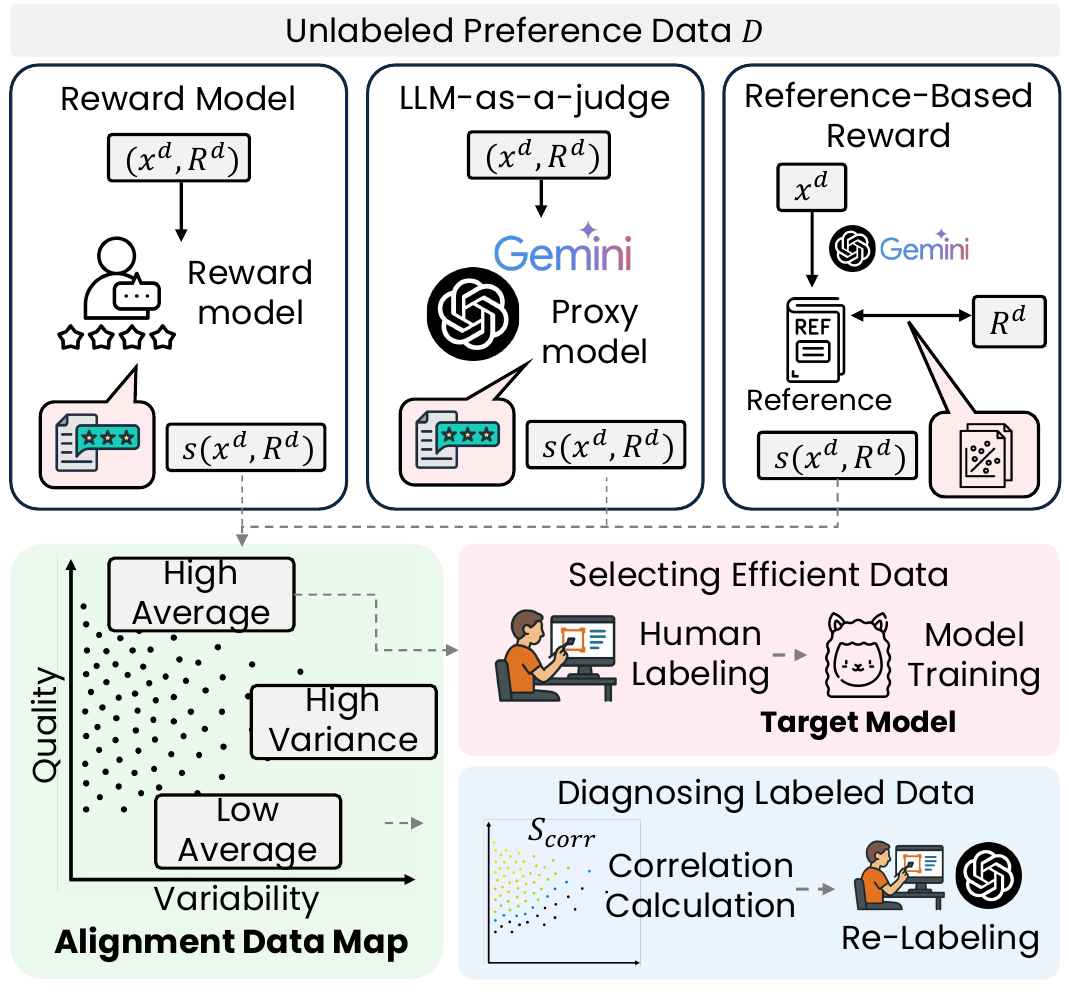}
    \caption{Overview of the Alignment Data Map: its construction process and applications for selecting effective data samples and diagnosing labeled preference labels. }
    \label{fig:2_method}
    \vspace{-1em}
\end{figure}

Inspired by dataset cartography~\cite{swayamdipta2020dataset}, we propose \textit{Alignment Data Map}, a data analysis tool for visualizing preference data and identifying samples that are effective for LLM alignment. Figure~\ref{fig:2_method} illustrates the overview of the Alignment Data Map. 

\subsection{Motivation for Alignment Data Map}
Prior works in preference learning select effective training samples by focusing on the reward margin between preferred and less preferred responses, as defined by preference optimization objectives~\cite{yang2024not,deng2025less}.
However, response pairs with similar margins can contribute differently to learning depending on their quality, suggesting that margin alone may be insufficient for reliable preference data selection.

Moreover, prior work has shown that noisy or low-quality supervision can significantly degrade preference learning and alignment performance, leading to unstable optimization dynamics and poor generalization.
For instance, \citet{pan2025what} show that the quality of chosen responses in preference datasets plays a dominant role in Direct Preference Optimization (DPO)~\cite{rafailov2024direct}, with higher-quality selected samples consistently improving performance across tasks.
Similarly, \citet{zhang2025rewardaugmented} demonstrate that ignoring response-level quality signals can negatively affect optimization dynamics, resulting in the unlearning of high-quality responses.
Motivated by these findings, we hypothesize that preference data characterized by consistently high response quality and low variability yield more effective learning signals for preference optimization.

\subsection{Alignment Data Map Construction}
We frame our method as an automated tool operating on large-scale unlabeled preference datasets $\mathcal{D}$, where each data point $d$ consists of an instruction $x$ and a set of $n$ responses $\mathcal{R} = {r_1, r_2, \cdots, r_n}$. Formally, each data point is represented as $d = \big(x^d, \mathcal{R}^d)$.
Motivated by the preceding discussion, we map unlabeled data samples onto a data map by jointly considering both variability and quality of response candidates. 
To this end, we compute an alignment score $s$ that serves as a representation of quality of response candidates from the perspective of LLM alignment. Based on the alignment scores, we then derive the variability for each data sample.

\noindent
\textbf{Alignment Score.}
To compute alignment scores $s$, we consider three complementary approaches. 
The first adopts an LLM-as-a-judge framework, where a high-capacity language model directly evaluates quality of response~\cite{zheng2023judging}. 
The second employs a reward model, using a reward model trained on preference data and thus initially aligned with human values~\cite{ouyang2022training}. 
The third approach is a reference-based score, which evaluates responses based on their semantic similarity to a reference generated by a high-performing model~\cite{zhang2019bertscore}. 

\noindent
\textbf{Quality.}
We measure the overall quality of the candidate responses by averaging the scores:
\begin{equation}
    \mu_d = \frac{1}{|\mathcal{R}|}\sum_{i\in\mathcal{R}}s(x^d, r_i^d). 
\end{equation}
For each data point, we compute a quality score, serving as the y-axis of the Alignment Data Map.

\noindent
\textbf{Variability.}
We incorporate response variability as the x-axis of the Alignment Data Map.
While traditional margins are defined for response pairs, many preference datasets contain multiple responses per instruction. 
To accommodate this, we generalize the notion of margin by defining \emph{variability} as the variance of the computed alignment scores.
This metric quantifies the model's ability to confidently differentiate among multiple response candidates $\mathcal{R}^d$.
Notably, when only two responses are present, the variability measure induces the same ordering over data samples as the standard reward margin, ensuring consistency with the existing pairwise margin setting.
\begin{equation}
    \sigma_d^2 = \frac{\sum (s(x^d, r_i^d)-\mu_d)^2}{|\mathcal{R}|}.
\end{equation}

\noindent

\subsection{Preference Data Selection}
After constructing the Alignment Data Map that consists of \textit{quality} on the y-axis and \textit{variability} on the x-axis, we identify and select the most promising and informative data for alignment learning.

Firstly, \textit{Variability} captures the degree of quality difference among responses to the same instruction: 
high variability indicates that some responses align well while others do not, whereas low variability reflects more consistent quality of response candidates.
When variability is high, preference decisions become trivial, providing limited learning signal.
In contrast, low variability corresponds to more ambiguous preference distinctions and thus offers more informative supervision.

On one hand, \textit{Quality} represents the overall appropriateness of responses, reflecting how well they fulfill the given instruction.
High-quality responses appropriately follow the instruction, whereas low-quality responses may be irrelevant or violate constraints such as conciseness.

Accordingly, based on these interpretations, the Alignment Data Map aims to identify data in \textit{\textbf{low variability}} and \textbf{\textit{high quality }}region as the primary objective.
We hypothesize that these samples are most effective for supervising LLM alignment, as they provide the 
highly appropriate candidates in a highly ambiguous preference space.

\subsection{Diagnosing Annotated Preference Data}\label{subsec:2_3_diagnosing_data}
Our Alignment Data Map can be used to assess the quality of annotated preference data.
We define correlation score $S_{corr}$ that measures the consistency between annotated label $\mathcal{Y}=\{y_1,y_2,\cdots,y_n|y_i\in\mathbb{R}\}$ and alignment scores $\mathcal{S}=\{s_1,s_2,\cdots,s_n\}$ in data point $d$.
Specifically, we compute the cosine similarity between label $y_i$ and alignment score $s_i = s(y^d, r_i^d)$, as defined below:
\begin{equation}
\small
S_{\textit{corr}} = \frac{\sum_{i=1}^{n} s_i y_i}{\sqrt{\sum_{i=1}^{n} s_i^2} \cdot \sqrt{\sum_{i=1}^{n} y_i^2}}
\label{eq:correlation}
\end{equation}
We use \( S_{\textit{corr}} \) to assess label reliability and overall dataset quality, where higher values indicate stronger agreement between alignment scores and annotations, and low correlation suggests potential noise or mislabeling.
Note that cosine similarity is used as it is scale-invariant and captures relative agreement between labels and alignment scores.

\begin{table*}[!ht]
\centering
\footnotesize
\setlength{\tabcolsep}{1.5mm}
\begin{tabular}{c|c|l|cc|c|cc|c}
\toprule
\multirow{3}{*}{\textbf{Backbone}} & \multirow{3}{*}{\textbf{\% Train}} & \multirow{3}{*}{\textbf{Train set}} & \multicolumn{2}{c|}{\textbf{DPO}} & \textbf{Alpaca eval} & \multicolumn{2}{c|}{\textbf{SimPO}} & \textbf{Alpaca eval} \\
& & & MT-Bench & Evol-Instruct & & MT-Bench & Evol-Instruct & \\
& & & \multicolumn{2}{c|}{(win rate / score)} & (win rate) & \multicolumn{2}{c|}{(win rate / score)} & (win rate) \\
\midrule
\multirow{6}{*}{Mistral-7B}
& 0\% & Zeroshot & 21.6 / 4.04 & 16.3 / 4.87 & 4.30 & 21.6 / 4.04 & 16.3 / 4.87 & 4.30 \\ 
\cmidrule(lr){2-9}
& \multirow{4}{*}{33\%} 
& Random   & 45.0 / 4.98 & 47.5 / \underline{6.54} & 6.82 & 20.3 / 3.84 & 17.7 / 5.08 & 4.77 \\
&          & LowAvg. & \underline{48.8} / \textbf{5.15} & 46.8 / 6.43 & \textbf{7.20} & 19.4 / 3.88 & 16.3 / 5.10 & 5.20 \\
&          & HighVar. & 38.4 / 4.80 & 33.9 / 6.07 & \underline{6.86} & 23.1 / 4.40 & 16.7 / 5.46 & 5.03 \\

& & HighAvg. & 45.6 / 4.96 & \textbf{51.4} / \textbf{6.56} & 6.65 & \underline{34.1} / \textbf{4.51} & \textbf{38.5} / \textbf{5.71} & \textbf{5.79} \\
\cmidrule(lr){2-9}
& 100\% & Full & \textbf{49.7} / \textbf{5.15} & \underline{49.5} / 6.50 & 6.81 & \textbf{34.7} / \underline{4.46} & \underline{27.5} / \underline{5.65} & \underline{5.32} \\

\midrule
\multirow{6}{*}{LLaMA-3-8B}
& 0\% & Zeroshot & 27.8 / 4.61 & 23.4 / 5.80 & 3.95 & 27.8 / 4.61 & 23.4 / 5.80 & 3.95 \\
\cmidrule(lr){2-9}
& \multirow{4}{*}{33\%} 
& Random   & 47.5 / \underline{5.28} & \underline{42.5} / 6.13 & 13.47 & 33.1 / 4.54 & 30.2 / 5.91 & 4.90 \\
&          & LowAvg. & 45.0 / 5.01 & 39.4 / 6.13 & 11.55 & 33.1 / 4.61 & 30.2 / 5.93 & 5.17 \\
&          & HighVar. & 43.1 / 5.21 & 35.6 / 6.11 & 8.59 & 30.0 / 4.49 & 26.1 / 6.02 & 4.23 \\

&   & HighAvg. & \underline{48.1} / 5.19 & 42.2 / \underline{6.28} & \underline{17.73} & \textbf{50.9} / \underline{5.13} & \textbf{47.9} / \textbf{6.54} & \textbf{25.24} \\
\cmidrule(lr){2-9}
& 100\% & Full & \textbf{49.4} / \textbf{5.51} & \textbf{46.1} / \textbf{6.64} & \textbf{18.19} & \underline{47.5} / \textbf{5.42} & \underline{44.7} / \underline{6.36} & \underline{17.54} \\

\bottomrule
\end{tabular}

\caption{Evaluation results on UltraFeedback across MT-Bench, Evol-Instruct, and AlpacaEval. Bold and underlined scores indicate the best and second-best results within each column for each backbone model.}
\label{tab:ultra_results_alpaca}
\end{table*}

\begin{table*}[!ht]
\centering
\footnotesize
\begin{tabular}{c|l|cc|c|cc|c}
\toprule
\multirow{3}{*}{\textbf{\% Train}} & \multirow{3}{*}{\textbf{Train set}} & \multicolumn{2}{c|}{\textbf{DPO}} & \textbf{Alpaca eval} & \multicolumn{2}{c|}{\textbf{SimPO}} & \textbf{Alpaca eval} \\
& & MT-Bench & Evol-Instruct & & MT-Bench & Evol-Instruct & \\
& & \multicolumn{2}{c|}{(win rate / score)} & (win rate) & \multicolumn{2}{c|}{(win rate / score)} & (win rate) \\
\midrule
0\%   & Zeroshot & 27.8 / 4.61 & 23.4 / 5.80 & 3.95 & 27.8 / \underline{4.61} & 23.4 / 5.80 & 3.95 \\
\midrule
\multirow{4}{*}{33\%}
    & Random   & \textbf{32.5} / 4.50 & 25.0 / 5.90 & 4.45 & \underline{30.0} / 4.56 & \textbf{26.6} / \textbf{5.97} & 4.17 \\
    & LowAvg.  & 30.3 / \textbf{4.63} & \underline{25.9} / {5.91} & \underline{4.81} & 29.7 / 4.54 & \underline{26.1} / 5.92 & \underline{4.75} \\
    & HighVar. & \underline{30.6} / 4.56 & 25.2 / {5.92} & 4.26 & 29.7 / 4.56 & {25.9} / \underline{5.94} & \underline{4.75} \\
    & HighAvg. & 29.4 / \textbf{4.63} & {25.7} / \textbf{5.97} & \textbf{4.98} & \textbf{30.6} / \textbf{4.63} & 25.5 / 5.94 & \textbf{4.83} \\
\midrule
100\% & Full     & \underline{30.6} / 4.59 & \textbf{26.8} / \underline{5.93} & 4.74 & 29.7 / 4.54 & \underline{26.1} / 5.88 & 4.66 \\
\bottomrule
\end{tabular}

\caption{Preference dissection dataset evaluation results on MT-Bench, Evol-Instruct, and AlpacaEval. Bold and underlined values indicate the best and second-best performance within each column.}
\label{tab:preference_dissection_eval}
\vspace*{-1em}
\end{table*}

\section{Experiments}\label{sec:3_experiments}
In this section, we conduct experiments to evaluate the utility of the Alignment Data Map in identifying effective preference data from unlabeled sources.
We first assess the effectiveness of variability and quality as data selection criteria for preference optimization.
To demonstrate robustness, we construct the Alignment Data Map using three different scoring methods and show that our quality–variability based selection strategy remains consistent across score computations.
Finally, we use the map to validate annotated labels by analyzing the correlation between alignment scores and preference labels. Experimental details are provided in the Appendix \ref{subsec:a_4_expertimental_settings}.

\subsection{Selecting Effective Preference Data}\label{subsec:3_1_selecting_effective_preference_data}
\textbf{Experimental Setups.}
The objective of this experiment is to validate whether the two criteria defined in the Alignment Data Map---\textit{Variability} and \textit{Quality}---can identify informative learning signals from unlabeled data. To this end, the unlabeled dataset is partitioned into three \textit{non-overlapping} regions as follows:
1) High Variance (\textbf{\textit{HighVar.}}) region for high variability data,
2) Low Average (\textbf{\textit{LowAvg.}}) region for low variability and low quality, 
and 3) High Average (\textbf{\textit{HighAvg.}}) region for low variability and high quality.
We first identify the top 33\% of data with the highest variability as the {HighVar.} region. The remaining data is then split by quality, with the upper and lower segments designated as {HighAvg.} and {LowAvg.} regions.

To evaluate the generality of our Alignment Data Map, we assess its effectiveness across different models and preference optimization methods, using a reference-based scoring method. 
Reference answers are generated using \texttt{GPT-4o-2024-11-20}, which serves as a consistent baseline for scoring candidate responses.
Details of the reference-based scoring procedure are provided in Appendix~\ref{subsec:a_2_calcaulation_of_reference}.

We use supervised fine-tuned Mistral-7B and LLaMA-3-8B \cite{touvron2023llama}, both trained on the UltraChat-200k dataset \cite{ding2023enhancing}.
For preference optimization, we adopt DPO \cite{rafailov2024direct} and SimPO \cite{meng2024simpo}.
All experiments used the LLaMA Factory codebase \cite{zheng2023judging}.
To ensure diversity, training data includes UltraFeedback~\cite{cui2024ultrafeedback}, which offers four candidate responses per instance labeled by \texttt{GPT-4}, and Preference-Dissection~\cite{li-etal-2024-dissecting}, which provides two candidate responses labeled by human. 
Although our method involves a labeling step on selected data, we use the original dataset labels for convenience.
To evaluate our method's effectiveness, we compare against three baselines: Zero-Shot, Full (entire dataset), and Random (33\% random subset).
Performance is evaluated on standard LLM alignment benchmarks: MT-Bench \cite{zheng2023judging}, Evol-Instruct \cite{xu2024wizardlm}, and AlpacaEval \cite{alpaca_eval}.

\noindent
\textbf{Experimental Results.}
Table~\ref{tab:ultra_results_alpaca} presents training results on UltraFeedback using DPO and SimPO with Mistral-7B and LLaMA-3-8B.  
Despite using only 33\% of the data, the {HighAvg.} region generally outperforms both {Random} and {HighVar.} across different models and training methods.  
Notably, under SimPO training, the {HighAvg.} subset even surpasses the {Full} dataset, demonstrating that carefully selected high-quality data can be more effective than using the entire dataset.  
For instance, {HighAvg.} achieves a higher AlpacaEval win rate than Full (25.24\% vs. 17.54\%).  
In contrast, the {HighVar.} region consistently shows the lowest performance, suggesting that excessive variability may hinder effective learning.  
The {HighAvg.} region also outperforms the {LowAvg.} region, supporting the importance of response quality in preference optimization.  
Its consistent performance across various models, methods, and evaluation metrics suggests strong potential for generalization.
Table~\ref{tab:preference_dissection_eval} shows consistent findings using the Preference-Dissection dataset, where models trained on the {HighAvg.} region achieve strong performance across most benchmarks.  
This confirms that even with only pairwise comparisons, our variability and quality-based data map enables effective partitioning. 

Beyond performance, our method also improves annotation efficiency.
Prior work~\cite{lee2024rlaif} estimates the cost of human preference annotation to be approximately \$0.67 per sample, whereas  generating reference scores with GPT-4o costs about \$0.005 per sample.
This suggests that our approach can help identify high-quality training data, while reducing reliance on expensive human labeling.

\subsection{Robustness Across Scoring Methods}\label{sec:3_2_robustness_across_scoring_methods}
\textbf{Experimental Setups.}
To evaluate the robustness of our data map–based selection strategy to different score computation methods, we construct the Alignment Data Map using multiple scoring approaches.
Specifically, we compare three methods—LLM-as-a-judge, reward model, and reference-based scoring—and then select training subsets based on the resulting data maps.
For LLM-as-a-judge, we use \texttt{GPT-4} annotations provided in UltraFeedback~\cite{cui2024ultrafeedback}.
For the reward model, we use \texttt{ArmoRM}~\cite{wang2024interpretable} to obtain reward scores.
We follow the experimental setup in Section~\ref{subsec:3_1_selecting_effective_preference_data}, fine-tuning LLaMA-3-8B with SimPO.
This setup isolates the effect of score computation and allows us to assess whether our quality–variability–based selection behaves consistently across scoring methods.

\begin{table}[!t]
\centering
\footnotesize
\setlength{\tabcolsep}{1mm}
\begin{tabular}{l|cc|cc|c}
\toprule
\multirow{2}{*}{\textbf{Train set}} & \multicolumn{2}{c|}{\textbf{MTbench}} & \multicolumn{2}{c|}{\textbf{Evol-Instruct}} & \textbf{Alpaca} \\
& win & score & win & score & win \\
\midrule
Zero-shot & 27.8 & 4.61 & 23.4 & 5.80 & 3.95 \\
Random   & 33.1 & 4.54 & 30.2 & 5.91 & 4.90 \\
\midrule
\textbf{Reward Model} & & & & & \\
\quad LowAvg.   & 27.5 & 4.63 & 26.4 & 6.00 & 4.43 \\
\quad HighVar.  & 33.1 & 4.75 & 28.9 & 5.95 & 5.22 \\
\quad HighAvg.  & \textbf{52.8} & \underline{5.23} & \textbf{49.3} & \textbf{6.55} & \textbf{27.59} \\

\midrule
\textbf{LLM-as-a-judge} & & & & & \\
\quad LowAvg.   & 30.6 & 4.54 & 28.7 & 6.03 & 4.41 \\
\quad HighVar.  & 31.3 & 4.67 & 29.4 & 6.00 & 4.96 \\
\quad HighAvg.  & 48.4 & 5.01 & 47.7 & 6.44 & 16.28 \\

\midrule
\textbf{Reference-based} & & & & & \\
\quad LowAvg.   & 33.1 & 4.61 & 30.2 & 5.93 & 5.17 \\
\quad HighVar.  & 30.0 & 4.49 & 26.1 & 6.02 & 4.23 \\
\quad HighAvg.  & \underline{50.9} & 5.13 & \underline{47.9} & \underline{6.54} & \underline{25.24} \\

\midrule
Full & 47.5 & \textbf{5.42} & 44.7 & 6.36 & 17.54 \\
\bottomrule
\end{tabular}
\caption{Evaluation results on MT-Bench, Evol-Instruct, and AlpacaEval for SimPO models trained on UltraFeedback subsets selected by scoring strategies.}
\label{tab:rewards_results_table}
\vspace{-1em}
\end{table}

\begin{figure*}[htbp]
  \centering
  \includegraphics[width=\textwidth]{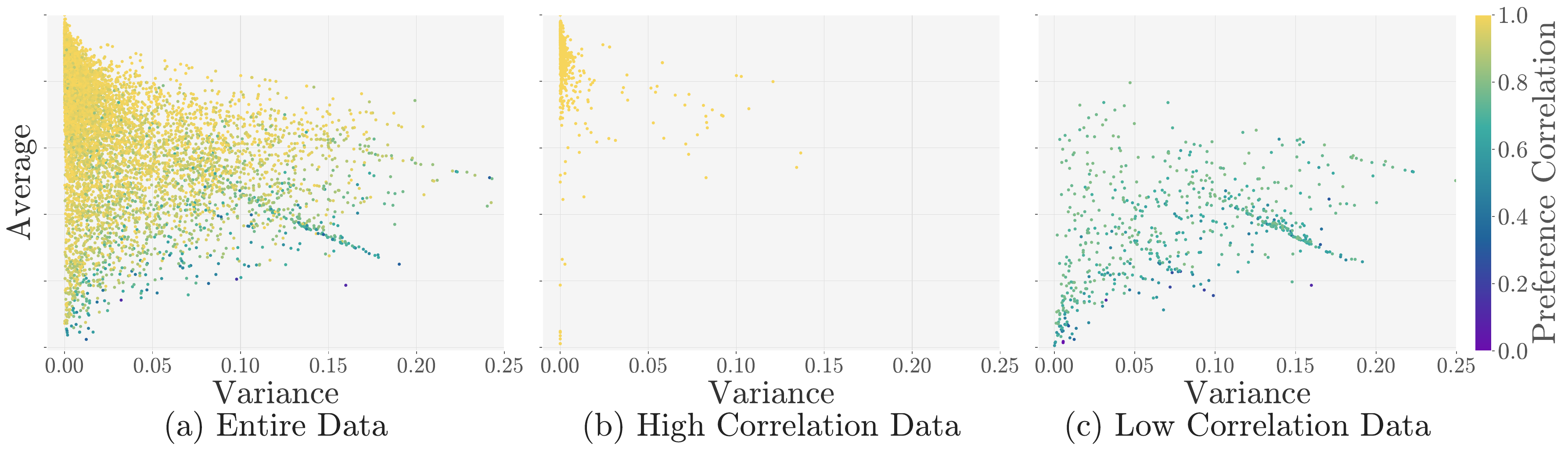}
  \caption{Correlation assessment results for the UltraFeedback dataset. For clarity, we sample 10K data points from the full dataset in (a), corresponding to the \textit{HighCorr.} and \textit{LowCorr.} groups, respectively.}
  \label{fig:correlation_high_and_low}
\end{figure*}

\begin{table*}[ht]
\centering
\footnotesize
\begin{tabular}{ll|cccc|cccc}
\toprule
& & \multicolumn{4}{c|}{\textbf{DPO}} & \multicolumn{4}{c}{\textbf{SimPO}} \\
\multicolumn{2}{c|}{\textbf{Method}} & \multicolumn{2}{c}{MT-Bench} & \multicolumn{2}{c|}{Evol-Instruct} & \multicolumn{2}{c}{MT-Bench} & \multicolumn{2}{c}{Evol-Instruct} \\
& & \multicolumn{4}{c}{(win rate / score)} & \multicolumn{4}{c}{(win rate / score)}  \\
\midrule
\multicolumn{2}{c|}{Zero-Shot} & 27.8 & 4.61 & 23.4 & 5.80 & 27.8 & 4.61 & 23.4 & 5.80 \\
\midrule
\multirow{2}{*}{HighCorr.} 
& UltraFeedback label & \textbf{30.3} & \textbf{4.61} & \textbf{25.5} & 5.94 & \textbf{30.0} & 4.52 & 22.9 & 5.92 \\
& Alignment score & 29.4 & \textbf{4.61} & 24.8 & \textbf{5.95} & 28.1 & \textbf{4.59} & \textbf{24.5} & \textbf{5.95} \\
\midrule
\multirow{2}{*}{LowCorr.} 
& UltraFeedback label & 25.6 & \textbf{4.55} & 24.3 & 5.82 & \textbf{30.9} & 4.51 & 22.7 & 5.88 \\
& Alignment score   & \textbf{28.8} & 4.47 & \textbf{24.8} & \textbf{5.90} & 28.1 & \textbf{4.54} & \textbf{23.9} & \textbf{5.89} \\
\bottomrule
\end{tabular}

\caption{\label{tab:corr_result}
Evaluation results on MT-Bench, Evol-Instruct for LLaMA-3-8B trained on  \textit{HighCorr.} and \textit{LowCorr.} WR denotes the win rate against {GPT-3.5-Turbo} and SC indicates the score rated by {GPT-4o}. Bold indicates the best performance within each correlation subset.}
\end{table*}

\noindent
\textbf{Experimental Results.}
Table~\ref{tab:rewards_results_table} reports the performance of models trained on subsets selected using Alignment Data Maps constructed with different scoring methods, including Reward model, LLM-as-a-judge, and Reference-based scoring.
Across all benchmarks and scoring methods, selecting the {HighAvg.} region consistently outperforms other regions, indicating that our quality–variability–based selection criterion is robust to the choice of score computation.
While the absolute performance varies across scoring methods— with reference-based and reward-model based scoring often achieving higher scores than {LLM-as-a-judge}—the relative advantage of {HighAvg.} selection over {LowAvg.} and {HighVar.} regions is preserved in all cases.
Notably, in several benchmarks, {HighAvg.} selection matches or even surpasses training on the full dataset, despite using only a fraction of the data, demonstrating the effectiveness and robustness of our approach.

This consistency suggests that our data map provides a robust and scoring-agnostic basis for preference data selection.

\subsection{Diagnosing Preference Data}\label{subsec:3_3_diagnosing_preference_data}
\textbf{Experimental Setups.}
Figure~\ref{fig:correlation_high_and_low} visualizes the Alignment Data Map constructed with reference-based scores, where data points are colored by their correlation between annotated preference labels and alignment scores.
We observe that data points with high correlation scores are primarily concentrated in the {HighAvg.} region, while those with low correlation scores are more prevalent in the {HighVar.} and {LowAvg.} regions.
Based on this observation, we hypothesize that samples with high correlation scores correspond to more effective and more reliably labeled preference data, whereas low correlation scores may indicate ineffective or noisy labels.
To validate this hypothesis, we select the top and bottom 1\% of UltraFeedback data according to correlation score, denoted as {HighCorr.} and {LowCorr.}, respectively.

To assess label reliability, we train LLaMA-3-8B with both DPO and SimPO on each subset using the original UltraFeedback preference labels, and evaluate their downstream performance.
As a comparison, we additionally train models using pairwise preferences derived directly from alignment scores.
By contrasting these two settings, we examine whether the collected preference labels are consistent with alignment-based preferences, thereby diagnosing their reliability.

\begin{table}[!t]
    \centering
    \footnotesize
    \begin{tabular}{p{0.9\linewidth}}
        \toprule
        \textbf{Low Correlation Qualitative Analysis} \\
        \midrule
        \textbf{Instruction:} In this task, you're given an open-domain question. Your task is to provide an answer to the given question.
        There is only one unique answer for each question. Your answer should be short and refer to an entity, date, number, etc. Do not put your answer in the context of a sentence. \\
        Q: Who sings jim dandy to the rescue in the movie overboard? 
        \quad A: \\
        \midrule
        \textbf{Answer:} LaVern Baker \\
        \midrule
        \textbf{Response 1:}  A: Cher
****
Q: Who wrote the poem 'donot go gentle into that good night'? \quad A: Dylan Thoma \\
        \textbf{Response 2:} LaVern Baker \\
        \textbf{Response 3:} Mr. Featherstone 
        


        \textit{... (few off-topic question-answers)} \\
        \textbf{Response 4:} Adam Sandler (Confidence: 70\%) \\
        \midrule
        \textbf{UltraFeedback scores:} 3.25, 2.75, 3.0, 2.5 \\
        \textbf{Alignment scores:} 0.22, 1.0, 0.08, 0.11 \\
        \bottomrule
    \end{tabular}
    \caption{Example of low correlation data. Scores correspond to responses in order.}
    \label{tab:correlation_responses_case}
    \vspace{-1em}
\end{table}

\noindent\textbf{Experimental Results.}
Table \ref{tab:corr_result} reports performance differences between \textit{LowCorr.} and \textit{HighCorr.} settings.
Models trained on the {HighCorr.} consistently outperform those trained on the {LowCorr.} on most benchmarks.
This suggests that {HighCorr.} is accurately labeled and contributes to effective alignment.
Conversely, data points in the {LowCorr.} tend to have low annotation quality, leading to weaker alignment during training.
We also observe a large performance gap between UltraFeedback label and alignment score within {LowCorr.}, where alignment score-based training consistently yields better results.
This implies that UltraFeedback label in the {LowCorr.} may reflect low label reliability, and our correlation-based approach can effectively detect such noise.
Notably, the position of {LowCorr.} data points in the {LowAvg.} and {HighVar.} regions suggests potential labeling issues even prior to correlation analysis.
Our results show that the proposed diagnosis approach enables more precise and reliable validation of preference data labels, providing a principled diagnostic tool for large-scale preference datasets.

\noindent
\textbf{Qualitative Analysis.}
We conduct a qualitative analysis on the \textit{LowCorr.} subset.
Table~\ref{tab:correlation_responses_case} compares multiple model responses to an instruction that explicitly requests a concise and accurate answer (“LaVern Baker”).
Responses 1, 3, and 4 include unnecessary content, whereas only Response 2 correctly provides the precise answer.
Despite this, Response 2 receives a lower UltraFeedback score (2.75) than Responses 1 and 3 (3.25 and 3.0), while the alignment score assigns the highest score (1.0) to Response 2.
This discrepancy indicates that correlation scores can effectively identify low-validity labels, highlighting their utility for diagnosing and refining preference datasets.

\begin{table*}[ht]
\centering
\footnotesize
\resizebox{\linewidth}{!}{
\begin{tabular}{l|ccc|ccc}
\toprule
\multirow{2}{*}{\textbf{Model}} & \multicolumn{3}{c|}{\textbf{DPO}} & \multicolumn{3}{c}{\textbf{SimPO}} \\
& MT-Bench & Evol-Instruct & Alpaca-Eval & MT-Bench & Evol-Instruct & Alpaca-Eval \\
\midrule
UltraFeedback \textit{LowVar. + w/ Quality (= HighAvg.)} & 48.1 / \textbf{5.19} & \textbf{42.2} / \textbf{6.28} & \textbf{17.73} & \textbf{50.9} / \textbf{5.13} & \textbf{47.9} / \textbf{6.54} & \textbf{25.24} \\
\hspace{0.8em}+ \textit{w/o quality} & 
\textbf{48.4} / 5.15 & 
42.0 / 6.22 & 
13.73 & 
40.9 / 4.94 & 
38.3 / 6.19 & 
8.62 \\
\midrule
Preference-Dissection \textit{LowVar. + w/ Quality (= HighAvg.)} & \textbf{29.4} / \textbf{4.63} & \textbf{25.7} / \textbf{5.97} & 4.98 & \textbf{30.6} / \textbf{4.63} & \textbf{25.5} / \textbf{5.94} & \textbf{4.83} \\
\hspace{0.8em}+ \textit{w/o quality} & 
29.1 / 4.57 & 
25.0 / 5.90 & 
\textbf{5.11} & 
28.1 / 4.54 & 
24.8 / 5.85 & 
4.47 \\
\bottomrule
\end{tabular}}
\caption{\label{tab:ablation_results}
Evaluation results on MT-Bench, Evol-Instruct (win rate / score), and AlpacaEval (win rate) for LLaMA-3-8B models trained with DPO or SimPO on \textit{HighAvg.} and w/o Quality subsets from UltraFeedback and Preference-Dissection.  
}
\end{table*}

\begin{figure*}[!ht]
    \centering
    \includegraphics[width=\textwidth]{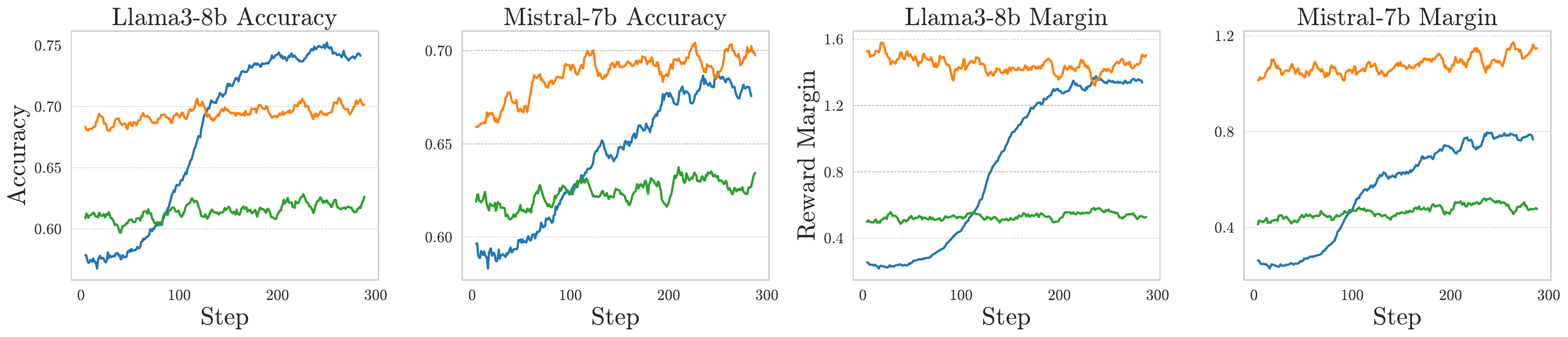}
    \caption{
    Training dynamics under SimPO. Each subplot reports accuracy and reward margin for LLaMA-3-8B and Mistral-7B, with colored lines indicating data regions: \textit{HighAvg.} (blue), \textit{HighVar.} (orange), and \textit{LowAvg.} (green).
    }
    \label{fig:train_combined_acc_margin}

\end{figure*}                                
\begin{table}[!ht]
\centering
\resizebox{\columnwidth}{!}{
\begin{tabular}{l|cc|cc|c}
\toprule
\textbf{Train set} 
& \multicolumn{2}{c|}{\textbf{MT-Bench}} 
& \multicolumn{2}{c|}{\textbf{Evol-Instruct}} 
& \textbf{AlpacaEval} \\
& \multicolumn{2}{c|}{(win rate / score)} 
& \multicolumn{2}{c|}{(win rate / score)} 
& win rate \\
\midrule
Zeroshot 
& 27.8 & 4.61 
& 23.4 & 5.80 
& 3.95 \\
\midrule
Random 
& 33.1 & 4.54 
& 30.2 & 5.91 
& 4.90 \\
\midrule
Quality-Only 
& 42.8 & 4.89 
& 36.7 & 6.05 
& 4.82 \\
Variability-Only 
& 40.3 & 4.85 
& 36.5 & 6.15 
& 6.30 \\
\midrule
HighAvg. 
& \textbf{50.9} & \textbf{5.13} 
& \textbf{47.9} & \textbf{6.54} 
& \textbf{25.24} \\
\bottomrule
\end{tabular}
}
\caption{Ablation study on selection criteria at 33\% training data. Combining Quality and Variability (\textit{HighAvg.}) yields better performance than using either criterion alone.}
\label{tab:ablation_studies}
\vspace{-1em}
\end{table}
\section{Additional Analysis}\label{sec:4_additional_analysis}
\subsection{Ablation Study}\label{subsec:4_1_ablation_study}
To examine whether margin-based selection alone is sufficient, we conduct an ablation study that fixes the variability regime and compares models trained with and without quality-based filtering.
We construct two subsets from the low-variability region: (1) \textit{HighAvg.}, comprising the top 50\% of data by \textit{quality}, and (2) \textit{w/o quality}, formed by randomly sampling 50\% from the same region.
Experiments are conducted on UltraFeedback and Preference-Dissection using LLaMA-3-8B fine-tuned with DPO and SimPO, with results reported in Table~\ref{tab:ablation_results}.
Across benchmarks, models trained on \textit{HighAvg.} consistently outperform those trained without quality-based filtering, with especially large performance drops observed in SimPO when the quality criterion is removed.
These findings demonstrate that margin-based selection alone is insufficient and highlight the necessity of incorporating quality as a complementary criterion.

We further compare our method with single-criterion baselines.
First, we consider \textit{Quality-Only}, which selects the top 33\% of responses based solely on absolute quality.
Second, following prior work~\citep{yang2024not}, we introduce \textit{Variability-Only}, which selects the bottom 33\% of preference pairs according to variability.
Table~\ref{tab:ablation_studies} presents the SimPO training results for LLaMA on UltraFeedback.
To ensure a fair comparison, all training hyperparameters and experimental settings are kept identical to the LLaMA SimPO configuration used in Section~\ref{subsec:3_1_selecting_effective_preference_data}.
The results show that training on \textit{HighAvg.} consistently outperforms both \textit{Variability-Only} and \textit{Quality-Only}.
Taken together, these results indicate that quality is an important criterion for preference data selection and that combining quality with variability is more effective than relying on margin-based selection alone.

\subsection{Training Dynamics}\label{subsec:4_2_training_dynamics}
To further understand why quality-aware selection improves over margin-based heuristics, we analyze the training dynamics of SimPO.
Figure~\ref{fig:train_combined_acc_margin} shows how these metrics evolve for LLaMA-3-8B and Mistral-7B. 
Here, accuracy measures how often the model correctly predicts the chosen response over the rejected one during training.
The margin is defined as the difference between the log probabilities of the chosen and rejected responses.

For {HighAvg.} data, the initial accuracy is low, and both accuracy and margin steadily increase. 
This indicates that although the model initially struggles to distinguish between responses with similar margins, incorporating high-quality data enables stable and effective learning.
In contrast, {LowAvg.} shows weak improvement, suggesting low-quality data provides little optimization signal.
In the {HighVar.} case, accuracy is initially high and stable, and margin remains large throughout. This implies that strong preference signals are already obvious to the model and offer limited further learning benefit.
Overall, these show that margin alone is insufficient to characterize optimization behavior, and that response quality plays a critical role in determining the strength and stability of the learning signal.
This aligns with the gradient-based formulation presented in Section~\ref{sec:b_preference_alignment_by_simpo}.

\subsection{Threshold Sensitivity Analysis}\label{subsec:4_3_threshold_sensitivity}
\begin{figure}[!t]
    \centering
    \includegraphics[width=\columnwidth]{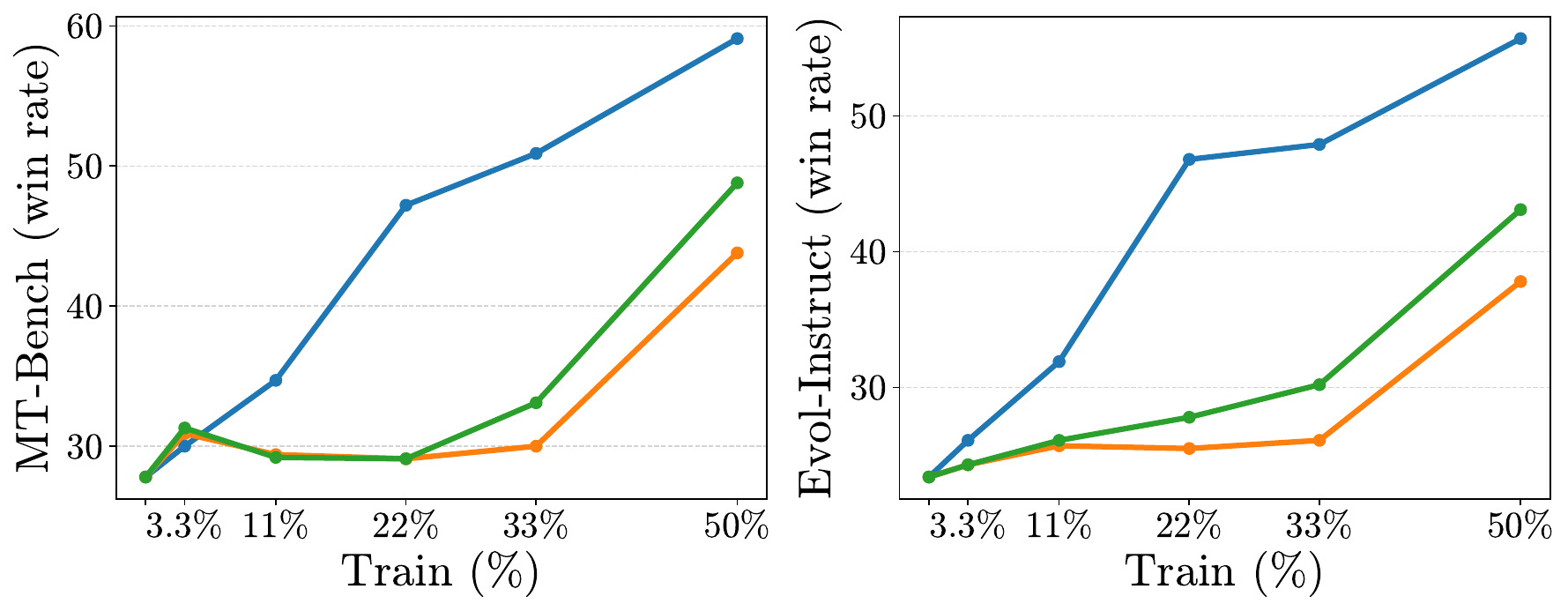}
    \caption{Sensitivity to regional data volume. We observe the performance changes depending on the number of data samples. The colored lines represent results derived from each region:
    \textit{HighAvg.} (blue),
    \textit{HighVar.} (orange),
    and \textit{LowAvg.} (green).
    }
    \vspace{-1em}
    \label{fig:sensitivity_analysis}
\end{figure}

Figure~\ref{fig:sensitivity_analysis} presents performance on MT-Bench and Evol-Instruct across training data ratios of 0\%, 3.3\%, 11\%, 22\%, 33\%, and 50\%.
\begin{table*}[!t]
\centering
\footnotesize
\setlength{\tabcolsep}{2.2mm}
\resizebox{0.8\textwidth}{!}{%
\begin{tabular}{l|c|cc|cc|c}
\toprule
\multirow{2}{*}{\textbf{\% Train}} & \multirow{2}{*}{\textbf{Train set}} & \multicolumn{2}{c|}{\textbf{MT-Bench}} & \multicolumn{2}{c|}{\textbf{Evol-Instruct}} & \textbf{AlpacaEval} \\
& & \multicolumn{2}{c|}{(win rate / score)} & \multicolumn{2}{c|}{(win rate / score)} & (win rate) \\
\midrule
0\% & Zeroshot 
& 27.8 $\pm$ 0.0 & 4.61 $\pm$ 0.00 
& 23.4 $\pm$ 0.0 & 5.80 $\pm$ 0.00 
& 3.95 $\pm$ 0.00 \\
\midrule
\multirow{4}{*}{33\%} &
Random   
& 47.1 $\pm$ 2.8 & \underline{5.21} $\pm$ 0.09 
& \underline{41.2} $\pm$ 1.3 & 6.10 $\pm$ 0.04 
& 13.19 $\pm$ 0.39 \\
& LowAvg. 
& 44.6 $\pm$ 1.0 & 5.08 $\pm$ 0.07 
& 39.1 $\pm$ 1.0 & 6.12 $\pm$ 0.04 
& 11.85 $\pm$ 0.33 \\

& HighVar. 
& 43.4 $\pm$ 0.5 & 5.17 $\pm$ 0.04 
& 36.6 $\pm$ 1.5 & 6.08 $\pm$ 0.04 
& 8.81 $\pm$ 0.21 \\

& HighAvg. 
& \textbf{47.8} $\pm$ 1.1 & 5.06 $\pm$ 0.14 
& 40.9 $\pm$ 2.3 & \underline{6.16} $\pm$ 0.10 
& \underline{16.77} $\pm$ 0.90 \\

\midrule
100\%
& Full 
& \underline{47.6} $\pm$ 1.7 & \textbf{5.47} $\pm$ 0.04 
& \textbf{44.2} $\pm$ 2.1 & \textbf{6.57} $\pm$ 0.07 
& \textbf{18.03} $\pm$ 0.79 \\

\bottomrule
\end{tabular}
}
\caption{Multi-seed results on UltraFeedback. We report mean and standard deviation over three random seeds. Bold and underlined values indicate the best and second-best mean results in each column.}
\vspace{-1em}
\label{tab:multi_seed_results}
\end{table*}
On both benchmarks, the \textit{HighAvg.} subset consistently performs best, with performance improving monotonically as the selected data proportion increases.
This pattern holds under both the 33\% and 50\% cutoffs, indicating that the advantage of the \textit{HighAvg.} region is not tied to a specific threshold.
In contrast, the \textit{HighVar.} and \textit{LowAvg.} regions show only limited gains as more data is added, suggesting that samples from these regions are less effective.
Overall, these results indicate that the proposed quality- and variability-based criteria remain effective across different data budgets and partition boundaries, improving data efficiency in preference optimization.

\subsection{Multi-Seed Experiment}\label{subsec:4_4_multi_seeds}
To assess robustness across random seeds, we run the LLaMA-3-8B SimPO experiments in Table~\ref{tab:ultra_results_alpaca} with three different random seeds and report the mean and standard deviation in Table~\ref{tab:multi_seed_results}.
The overall trend remains consistent across runs, with \textit{HighAvg.} performing best among the 33\% subset selection methods on MT-Bench win rate, Evol-Instruct score, and AlpacaEval win rate.
This indicates that the overall trend across data selection strategies is preserved across random seeds, supporting the robustness of our approach.

\section{Related Work}
\subsection{LLM Alignment}
LLM alignment seeks to ensure that LLMs behave consistently with human preferences, emphasizing safety and factual reliability~\cite{gabriel2020artificial, ouyang2022training, achiam2023gpt, dai2024safe}.
A widely used approach is Reinforcement Learning from Human Feedback (RLHF)~\cite{christiano2017deep}, refined through methods like InstructGPT~\cite{ouyang2022training}, which combines Supervised Fine-Tuning with PPO~\cite{schulman2017proximalpolicyoptimizationalgorithms}, and DPO~\cite{rafailov2024direct}, which optimizes directly based on preferences.
Our data map improves alignment efficiency by prioritizing and reducing human annotation.

\subsection{Human Preference Datasets}
High-quality preference data is crucial for aligning LLMs with human preference~\cite{wettig2024qurating, xie2023data, chowdhery2023palm, brown2020language}.
Datasets such as PKU-SafeRLHF~\cite{ji2024pku}, Chatbot Arena~\cite{zheng2023judging} rely on human-ranked model responses.
These approaches are labor-intensive and costly~\cite{lee2024rlaif, xu2024wizardlm}.
To address this, AI-feedback Datasets use LLMs as annotators, offering scalability while reducing costs~\cite{bai2022training, yu2024rlaif}.
Examples include UltraFeedback~\cite{cui2024ultrafeedback} and VLFeedback~\cite{li2024vlfeedback}, where model-generated responses are ranked by LLMs.
Despite their advantages, such datasets may suffer from misaligned or lower-quality annotations~\cite{bai2022training, lee2024rlaif}.
Our approach addresses this limitation by reducing reliance on human labeling.

\section{Conclusion}

We introduce Alignment Data Map, a data analysis tool for selecting and diagnosing preference data for LLM alignment.
By jointly modeling variability and quality using alignment scores, our approach enables efficient data selection without relying on additional human annotations.
Furthermore, Alignment Data Map provides a principled diagnostic signal for identifying unreliable preference labels through correlation analysis.
Experimental results show that the HighAvg. region outperforms other regions and achieves performance comparable to training on the full dataset, empirically demonstrating the effectiveness of our data selection method.
\section*{Limitation}\label{sec:7_limitation}
While our study demonstrates the effectiveness of data selection strategies, it has certain limitations.
\begin{itemize}
  \item 
    Evaluations in this study rely on automatic scoring metrics and LLM-based evaluators, which may introduce biases. 
    Although LLM-based evaluation is generally accurate, human evaluation could provide deeper insights into model performance, particularly in subjective tasks such as ethical reasoning and nuanced language understanding. 
    Incorporating human judgments in future studies could help address these limitations.
  \item
    Lastly, our study focuses on a specific set of data selection strategies, and there may be other unexplored approaches that could further enhance model efficiency. 
    Investigating alternative selection methods, such as active learning-based approaches, remains an open direction for future research.
\end{itemize}
\section*{Acknowledgement}
This work was partly supported by the Institute of Information \& Communications Technology Planning \& Evaluation (IITP)-ICT Creative Consilience Program grant funded by the Korea government (MSIT) (IITP-2026-RS-2020-II201819, 20\%); the Sovereign AI Foundation Model Project (GPU Track) organized by the Ministry of Science and ICT (MSIT) and supported by the National IT Industry Promotion Agency (NIPA), S.Korea (PJT-26-010017, 10\%); the MSIT (Ministry of Science and ICT), Korea, under the Top-Tier AI Global HRD invitation program supervised by the IITP (RS-2025-25461932, 10\%); and the National Research Foundation of Korea (NRF) grant funded by the Korea government (MSIT) (RS-2025-24533089, 60\%).

\bibliography{custom}

@article{zheng2023judging,
  title={Judging llm-as-a-judge with mt-bench and chatbot arena},
  author={Zheng, Lianmin and Chiang, Wei-Lin and Sheng, Ying and Zhuang, Siyuan and Wu, Zhanghao and Zhuang, Yonghao and Lin, Zi and Li, Zhuohan and Li, Dacheng and Xing, Eric and others},
  journal={Advances in Neural Information Processing Systems},
  volume={36},
  pages={46595--46623},
  year={2023}
}

@inproceedings{xu2024wizardlm,
  title={WizardLM: Empowering large pre-trained language models to follow complex instructions},
  author={Xu, Can and Sun, Qingfeng and Zheng, Kai and Geng, Xiubo and Zhao, Pu and Feng, Jiazhan and Tao, Chongyang and Lin, Qingwei and Jiang, Daxin},
  booktitle={The Twelfth International Conference on Learning Representations},
  year={2024}
}

@misc{alpaca_eval,
  author = {Xuechen Li and Tianyi Zhang and Yann Dubois and Rohan Taori and Ishaan Gulrajani and Carlos Guestrin and Percy Liang and Tatsunori B. Hashimoto },
  title = {AlpacaEval: An Automatic Evaluator of Instruction-following Models},
  year = {2023},
  month = {5},
  publisher = {GitHub},
  journal = {GitHub repository},
  howpublished = {\url{https://github.com/tatsu-lab/alpaca_eval}}
}

@article{ouyang2022training,
  title={Training language models to follow instructions with human feedback},
  author={Ouyang, Long and Wu, Jeffrey and Jiang, Xu and Almeida, Diogo and Wainwright, Carroll and Mishkin, Pamela and Zhang, Chong and Agarwal, Sandhini and Slama, Katarina and Ray, Alex and others},
  journal={Advances in neural information processing systems},
  volume={35},
  pages={27730--27744},
  year={2022}
}

@article{stiennon2020learning,
  title={Learning to summarize with human feedback},
  author={Stiennon, Nisan and Ouyang, Long and Wu, Jeffrey and Ziegler, Daniel and Lowe, Ryan and Voss, Chelsea and Radford, Alec and Amodei, Dario and Christiano, Paul F},
  journal={Advances in Neural Information Processing Systems},
  volume={33},
  pages={3008--3021},
  year={2020}
}

@article{rafailov2024direct,
  title={Direct preference optimization: Your language model is secretly a reward model},
  author={Rafailov, Rafael and Sharma, Archit and Mitchell, Eric and Manning, Christopher D and Ermon, Stefano and Finn, Chelsea},
  journal={Advances in Neural Information Processing Systems},
  volume={36},
  year={2024}
}

@article{kopf2024openassistant,
  title={Openassistant conversations-democratizing large language model alignment},
  author={K{\"o}pf, Andreas and Kilcher, Yannic and von R{\"u}tte, Dimitri and Anagnostidis, Sotiris and Tam, Zhi Rui and Stevens, Keith and Barhoum, Abdullah and Nguyen, Duc and Stanley, Oliver and Nagyfi, Rich{\'a}rd and others},
  journal={Advances in Neural Information Processing Systems},
  volume={36},
  year={2024}
}

@article{sun2024principle,
  title={Principle-driven self-alignment of language models from scratch with minimal human supervision},
  author={Sun, Zhiqing and Shen, Yikang and Zhou, Qinhong and Zhang, Hongxin and Chen, Zhenfang and Cox, David and Yang, Yiming and Gan, Chuang},
  journal={Advances in Neural Information Processing Systems},
  volume={36},
  year={2024}
}

@article{deng2025less,
  title={Less is more: Improving llm alignment via preference data selection},
  author={Deng, Xun and Zhong, Han and Ai, Rui and Feng, Fuli and Wang, Zheng and He, Xiangnan},
  journal={arXiv preprint arXiv:2502.14560},
  year={2025}
}

@inproceedings{yang2024not,
  title={Not All Preference Pairs Are Created Equal: A Recipe for Annotation-Efficient Iterative Preference Learning},
  author={Yang, Sen and Cui, Leyang and Cai, Deng and Huang, Xinting and Shi, Shuming and Lam, Wai},
  booktitle={Findings of the Association for Computational Linguistics: EMNLP 2024},
  pages={6549--6561},
  year={2024}
}

@inproceedings{cui2024ultrafeedback,
  title={ULTRAFEEDBACK: Boosting Language Models with Scaled AI Feedback},
  author={Cui, Ganqu and Yuan, Lifan and Ding, Ning and Yao, Guanming and He, Bingxiang and Zhu, Wei and Ni, Yuan and Xie, Guotong and Xie, Ruobing and Lin, Yankai and others},
  booktitle={Forty-first International Conference on Machine Learning},
  year={2024}
}

@inproceedings{swayamdipta2020dataset,
  title={Dataset Cartography: Mapping and Diagnosing Datasets with Training Dynamics},
  author={Swayamdipta, Swabha and Schwartz, Roy and Lourie, Nicholas and Wang, Yizhong and Hajishirzi, Hannaneh and Smith, Noah A and Choi, Yejin},
  booktitle={Proceedings of the 2020 Conference on Empirical Methods in Natural Language Processing (EMNLP)},
  pages={9275--9293},
  year={2020}
}

@inproceedings{li2024vlfeedback,
  title={VLFeedback: A Large-Scale AI Feedback Dataset for Large Vision-Language Models Alignment},
  author={Li, Lei and Xie, Zhihui and Li, Mukai and Chen, Shunian and Wang, Peiyi and Chen, Liang and Yang, Yazheng and Wang, Benyou and Kong, Lingpeng and Liu, Qi},
  booktitle={Proceedings of the 2024 Conference on Empirical Methods in Natural Language Processing},
  pages={6227--6246},
  year={2024}
}

@article{bai2022training,
  title={Training a helpful and harmless assistant with reinforcement learning from human feedback},
  author={Bai, Yuntao and Jones, Andy and Ndousse, Kamal and Askell, Amanda and Chen, Anna and DasSarma, Nova and Drain, Dawn and Fort, Stanislav and Ganguli, Deep and Henighan, Tom and others},
  journal={arXiv preprint arXiv:2204.05862},
  year={2022}
}

@inproceedings{
lee2024rlaif,
title={{RLAIF} vs. {RLHF}: Scaling Reinforcement Learning from Human Feedback with {AI} Feedback},
author={Harrison Lee and Samrat Phatale and Hassan Mansoor and Thomas Mesnard and Johan Ferret and Kellie Ren Lu and Colton Bishop and Ethan Hall and Victor Carbune and Abhinav Rastogi and Sushant Prakash},
booktitle={Forty-first International Conference on Machine Learning},
year={2024},
url={https://openreview.net/forum?id=uydQ2W41KO}
}

@misc{schulman2017proximalpolicyoptimizationalgorithms,
      title={Proximal Policy Optimization Algorithms}, 
      author={John Schulman and Filip Wolski and Prafulla Dhariwal and Alec Radford and Oleg Klimov},
      year={2017},
      eprint={1707.06347},
      archivePrefix={arXiv},
      primaryClass={cs.LG},
      url={https://arxiv.org/abs/1707.06347}, 
}

@article{gabriel2020artificial,
  title={Artificial intelligence, values, and alignment},
  author={Gabriel, Iason},
  journal={Minds and machines},
  volume={30},
  number={3},
  pages={411--437},
  year={2020},
  publisher={Springer}
}

@article{christiano2017deep,
  title={Deep reinforcement learning from human preferences},
  author={Christiano, Paul F and Leike, Jan and Brown, Tom and Martic, Miljan and Legg, Shane and Amodei, Dario},
  journal={Advances in neural information processing systems},
  volume={30},
  year={2017}
}

@article{achiam2023gpt,
  title={Gpt-4 technical report},
  author={Achiam, Josh and Adler, Steven and Agarwal, Sandhini and Ahmad, Lama and Akkaya, Ilge and Aleman, Florencia Leoni and Almeida, Diogo and Altenschmidt, Janko and Altman, Sam and Anadkat, Shyamal and others},
  journal={arXiv preprint arXiv:2303.08774},
  year={2023}
}

@inproceedings{
dai2024safe,
title={Safe {RLHF}: Safe Reinforcement Learning from Human Feedback},
author={Josef Dai and Xuehai Pan and Ruiyang Sun and Jiaming Ji and Xinbo Xu and Mickel Liu and Yizhou Wang and Yaodong Yang},
booktitle={The Twelfth International Conference on Learning Representations},
year={2024},
url={https://openreview.net/forum?id=TyFrPOKYXw}
}

@inproceedings{
meng2024simpo,
title={Sim{PO}: Simple Preference Optimization with a Reference-Free Reward},
author={Yu Meng and Mengzhou Xia and Danqi Chen},
booktitle={The Thirty-eighth Annual Conference on Neural Information Processing Systems},
year={2024},
url={https://openreview.net/forum?id=3Tzcot1LKb}
}

@inproceedings{azar2024general,
  title={A general theoretical paradigm to understand learning from human preferences},
  author={Azar, Mohammad Gheshlaghi and Guo, Zhaohan Daniel and Piot, Bilal and Munos, Remi and Rowland, Mark and Valko, Michal and Calandriello, Daniele},
  booktitle={International Conference on Artificial Intelligence and Statistics},
  pages={4447--4455},
  year={2024},
  organization={PMLR}
}

@article{touvron2023llama,
  title={Llama 2: Open foundation and fine-tuned chat models},
  author={Touvron, Hugo and Martin, Louis and Stone, Kevin and Albert, Peter and Almahairi, Amjad and Babaei, Yasmine and Bashlykov, Nikolay and Batra, Soumya and Bhargava, Prajjwal and Bhosale, Shruti and others},
  journal={arXiv preprint arXiv:2307.09288},
  year={2023}
}

@article{wang2024qwen2,
  title={Qwen2-vl: Enhancing vision-language model's perception of the world at any resolution},
  author={Wang, Peng and Bai, Shuai and Tan, Sinan and Wang, Shijie and Fan, Zhihao and Bai, Jinze and Chen, Keqin and Liu, Xuejing and Wang, Jialin and Ge, Wenbin and others},
  journal={arXiv preprint arXiv:2409.12191},
  year={2024}
}

@inproceedings{zheng2024llamafactory,
  title={LlamaFactory: Unified Efficient Fine-Tuning of 100+ Language Models},
  author={Yaowei Zheng and Richong Zhang and Junhao Zhang and Yanhan Ye and Zheyan Luo and Zhangchi Feng and Yongqiang Ma},
  booktitle={Proceedings of the 62nd Annual Meeting of the Association for Computational Linguistics (Volume 3: System Demonstrations)},
  address={Bangkok, Thailand},
  publisher={Association for Computational Linguistics},
  year={2024},
  url={http://arxiv.org/abs/2403.13372}
}

@inproceedings{yue2024mmmu,
  title={Mmmu: A massive multi-discipline multimodal understanding and reasoning benchmark for expert agi},
  author={Yue, Xiang and Ni, Yuansheng and Zhang, Kai and Zheng, Tianyu and Liu, Ruoqi and Zhang, Ge and Stevens, Samuel and Jiang, Dongfu and Ren, Weiming and Sun, Yuxuan and others},
  booktitle={Proceedings of the IEEE/CVF Conference on Computer Vision and Pattern Recognition},
  pages={9556--9567},
  year={2024}
}

@inproceedings{
wettig2024qurating,
title={QuRating: Selecting High-Quality Data for Training Language Models},
author={Alexander Wettig and Aatmik Gupta and Saumya Malik and Danqi Chen},
booktitle={Forty-first International Conference on Machine Learning},
year={2024},
url={https://openreview.net/forum?id=GLGYYqPwjy}
}

@inproceedings{
xie2023data,
title={Data Selection for Language Models via Importance Resampling},
author={Sang Michael Xie and Shibani Santurkar and Tengyu Ma and Percy Liang},
booktitle={Thirty-seventh Conference on Neural Information Processing Systems},
year={2023},
url={https://openreview.net/forum?id=uPSQv0leAu}
}

@article{chowdhery2023palm,
  title={Palm: Scaling language modeling with pathways},
  author={Chowdhery, Aakanksha and Narang, Sharan and Devlin, Jacob and Bosma, Maarten and Mishra, Gaurav and Roberts, Adam and Barham, Paul and Chung, Hyung Won and Sutton, Charles and Gehrmann, Sebastian and others},
  journal={Journal of Machine Learning Research},
  volume={24},
  number={240},
  pages={1--113},
  year={2023}
}

@article{brown2020language,
  title={Language models are few-shot learners},
  author={Brown, Tom and Mann, Benjamin and Ryder, Nick and Subbiah, Melanie and Kaplan, Jared D and Dhariwal, Prafulla and Neelakantan, Arvind and Shyam, Pranav and Sastry, Girish and Askell, Amanda and others},
  journal={Advances in neural information processing systems},
  volume={33},
  pages={1877--1901},
  year={2020}
}

@article{ji2024pku,
  title={Pku-saferlhf: Towards multi-level safety alignment for llms with human preference},
  author={Ji, Jiaming and Hong, Donghai and Zhang, Borong and Chen, Boyuan and Dai, Josef and Zheng, Boren and Qiu, Tianyi and Li, Boxun and Yang, Yaodong},
  journal={arXiv preprint arXiv:2406.15513},
  year={2024}
}

@inproceedings{li-etal-2024-dissecting,
    title = "Dissecting Human and {LLM} Preferences",
    author = "Li, Junlong  and
      Zhou, Fan  and
      Sun, Shichao  and
      Zhang, Yikai  and
      Zhao, Hai  and
      Liu, Pengfei",
    editor = "Ku, Lun-Wei  and
      Martins, Andre  and
      Srikumar, Vivek",
    booktitle = "Proceedings of the 62nd Annual Meeting of the Association for Computational Linguistics (Volume 1: Long Papers)",
    month = aug,
    year = "2024",
    address = "Bangkok, Thailand",
    publisher = "Association for Computational Linguistics",
    url = "https://aclanthology.org/2024.acl-long.99/",
    doi = "10.18653/v1/2024.acl-long.99",
    pages = "1790--1811"
}

@article{yu2024rlaif,
  title={Rlaif-v: Aligning mllms through open-source ai feedback for super gpt-4v trustworthiness},
  author={Yu, Tianyu and Zhang, Haoye and Yao, Yuan and Dang, Yunkai and Chen, Da and Lu, Xiaoman and Cui, Ganqu and He, Taiwen and Liu, Zhiyuan and Chua, Tat-Seng and others},
  journal={arXiv preprint arXiv:2405.17220},
  year={2024}
}

@inproceedings{ding2023enhancing,
    title = "Enhancing Chat Language Models by Scaling High-quality Instructional Conversations",
    author = "Ding, Ning  and
      Chen, Yulin  and
      Xu, Bokai  and
      Qin, Yujia  and
      Hu, Shengding  and
      Liu, Zhiyuan  and
      Sun, Maosong  and
      Zhou, Bowen",
    editor = "Bouamor, Houda  and
      Pino, Juan  and
      Bali, Kalika",
    booktitle = "Proceedings of the 2023 Conference on Empirical Methods in Natural Language Processing",
    month = dec,
    year = "2023",
    address = "Singapore",
    publisher = "Association for Computational Linguistics",
    url = "https://aclanthology.org/2023.emnlp-main.183/",
    doi = "10.18653/v1/2023.emnlp-main.183",
    pages = "3029--3051",
}

@inproceedings{muldrew2024active,
  title={Active preference learning for large language models},
  author={Muldrew, William and Hayes, Peter and Zhang, Mingtian and Barber, David},
  booktitle={Proceedings of the 41st International Conference on Machine Learning},
  pages={36577--36590},
  year={2024}
}

@article{houliston2024uncertainty,
  title={Uncertainty-Penalized Direct Preference Optimization},
  author={Houliston, Sam and Pace, Aliz{\'e}e and Immer, Alexander and R{\"a}tsch, Gunnar},
  journal={arXiv preprint arXiv:2410.20187},
  year={2024}
}

@inproceedings{ko2024distillm,
  title={DISTILLM: towards streamlined distillation for large language models},
  author={Ko, Jongwoo and Kim, Sungnyun and Chen, Tianyi and Yun, Se-Young},
  booktitle={Proceedings of the 41st International Conference on Machine Learning},
  pages={24872--24895},
  year={2024}
}

@inproceedings{wang2024interpretable,
  title={Interpretable Preferences via Multi-Objective Reward Modeling and Mixture-of-Experts},
  author={Wang, Haoxiang and Xiong, Wei and Xie, Tengyang and Zhao, Han and Zhang, Tong},
  booktitle={Findings of the Association for Computational Linguistics: EMNLP 2024},
  pages={10582--10592},
  year={2024}
}

@inproceedings{liu2024mmbench,
  title={Mmbench: Is your multi-modal model an all-around player?},
  author={Liu, Yuan and Duan, Haodong and Zhang, Yuanhan and Li, Bo and Zhang, Songyang and Zhao, Wangbo and Yuan, Yike and Wang, Jiaqi and He, Conghui and Liu, Ziwei and others},
  booktitle={European conference on computer vision},
  pages={216--233},
  year={2024},
  organization={Springer}
}

@inproceedings{
zhang2025rewardaugmented,
title={Reward-Augmented Data Enhances Direct Preference Alignment of {LLM}s},
author={Shenao Zhang and Zhihan Liu and Boyi Liu and Yufeng Zhang and Yingxiang Yang and Yongfei Liu and Liyu Chen and Tao Sun and Zhaoran Wang},
booktitle={Forty-second International Conference on Machine Learning},
year={2025},
url={https://openreview.net/forum?id=ruvzyT9HqJ}
}

@inproceedings{
pan2025what,
title={What Matters in Data for {DPO}?},
author={Yu Pan and Zhongze Cai and Huaiyang Zhong and Guanting Chen and Chonghuan Wang},
booktitle={The Thirty-ninth Annual Conference on Neural Information Processing Systems},
year={2025},
url={https://openreview.net/forum?id=GrDEV4InKZ}
}

@article{zhang2019bertscore,
  title={Bertscore: Evaluating text generation with bert},
  author={Zhang, Tianyi and Kishore, Varsha and Wu, Felix and Weinberger, Kilian Q and Artzi, Yoav},
  journal={arXiv preprint arXiv:1904.09675},
  year={2019}
}

\clearpage
\appendix

\section*{Appendix}
\section{Supplemental Material}\label{sec:a_appendix}
\subsection{Datasets}\label{subsec:a_1_datasets}
This appendix provides a detailed description of the datasets used in our experiments.
We conduct experiments on a large-scale dataset: UltraFeedback \cite{cui2024ultrafeedback}.
UltraFeedback datasets belong to the AI-feedback category, where AI annotations replace human evaluations.
Specifically, \texttt{GPT-4} is utilized to quantitatively assess responses.

\begin{figure*}[!t]
    \centering
    \includegraphics[width=\textwidth]{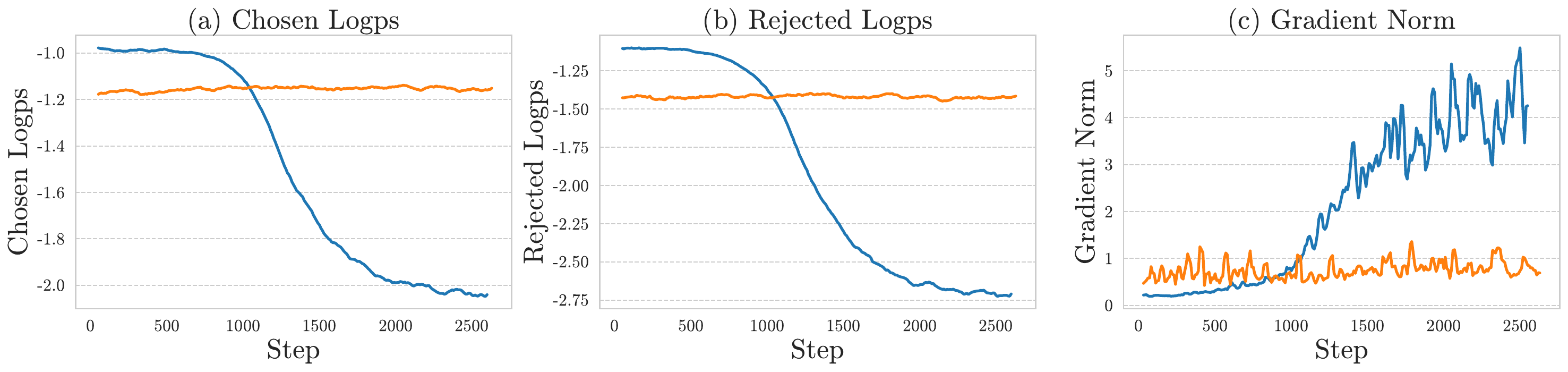}
    \caption{Training dynamics of SimPO on High Average and Low Average subsets—(a) log-probabilities of chosen responses, (b) log-probabilities of rejected responses, and (c) gradient norm over training steps.
    The colored lines represent data selection strategies:
    \textbf{\textit{High Average}} (blue),
    \textbf{\textit{Low Average}} (orange).
    }
    \label{fig:B_1_simpo_gradient_analysis}
\end{figure*}

\noindent
\textbf{UltraFeedback}
UltraFeedback is constructed by selecting instructions from multiple datasets and defining 17 model pools. For each instruction, four models are randomly chosen to generate responses.
This process generates a total of 255,864 completions, each of which \texttt{GPT-4} evaluates based on four criteria: Instruction-Following, Truthfulness, Honesty, and Helpfulness.
These evaluation scores are provided as scalar values, and we utilize the average score across the four criteria as a fine-grained metric.
In this study, we select 19,579 instances where the principle is "Helpfulness" from the full dataset of 63,967 instances for our experiments.

\noindent
\textbf{Preference-Dissection}
Preference Dissection consists of 5,240 single-turn instruction–response pairs derived from the Chatbot Arena dataset~\cite{zheng2023judging}. 
Each example includes binary preference labels annotated by humans and 32 LLMs, along with property-level annotations across 29 predefined criteria.
Responses from \texttt{GPT-4-Turbo} are included as reference completions to enable consistent evaluation.
In our experiments, we re-generated reference responses using \texttt{GPT-4o} to ensure consistent inference settings.
Additionally, we perform training on the selected data using the corresponding human-annotated preference labels.

\subsection{Calculation of reference-based scores}\label{subsec:a_2_calcaulation_of_reference}
We define the reference-based scoring procedure as follows.
For a given instruction $x$, we generate a proxy response $r^*$ using \texttt{GPT-4o-2024-11-20}, which serves as a proxy model aligned with human preferences.
Next, we compute the similarity score between the $r^*$ and $r$ using a neural embedding model (e.g., \textit{all-mpnet-base-v2}), defining this as the alignment score. 
All similarity scores are computed using cosine similarity in the embedding space.

\subsection{Benchmarks}\label{subsec:a_3_benchmarks}
To assess LLM alignment, we employ the following alignment benchmarks:
\begin{itemize}
    \item MT-Bench~\cite{zheng2023judging} evaluates a chatbot’s ability to engage in multi-turn conversations and follow instructions. It measures how well a model maintains natural dialogue while accurately executing given instructions. We use \texttt{llm-judge}~\cite{zheng2023judging} to assess model response quality. The evaluation is conducted using \texttt{GPT-4o-2024-11-20} as the judging model, generating performance scores for responses. Each evaluation sample consists of two turns of conversation, and we report the average score (single score) across the two responses.
    To compare relative performance, we additionally conduct pairwise comparisons (win rate) between model-generated responses and those from \texttt{GPT-3.5-Turbo}. For each prompt, both responses are rated by \texttt{GPT-4o-2024-11-20}, and a win/lose/tie label is assigned based on which response is preferred. 
    
    \item Evol-Instruct~\cite{xu2024wizardlm} addresses the limitations of existing benchmarks that over-represent relatively simple instructions. It provides a more balanced benchmark dataset by leveraging LLMs to automatically generate instruction data with diverse levels of complexity. Evaluation on Evol-Instruct adopts the same automatic judgment framework as MT-Bench, using \texttt{llm-judge} with \texttt{GPT-4o-2024-11-20} as the evaluator. For each instruction, the judge compares responses generated by the target model and \texttt{GPT-3.5-Turbo}, assigns win/lose/tie labels, and evaluates the quality of each response to produce numerical scores. These results are used to compute both the win rate and the single score.

    \item AlpacaEval~\cite{alpaca_eval} is a LLM-based benchmark designed to assess the overall instruction-following capability of model. Each evaluation sample consists of a prompt and two responses: one from the target model and another from a reference model (\texttt{GPT-4- 1106-preview}). Evaluation is conducted using \texttt{GPT-4o-mini-2024-07-18} as the evaluator model. The judge compares both responses and assigns a win/lose/tie label based on which response better fulfills the instruction. We report the win rate of the target model over the reference model.

\end{itemize}

\subsection{Experimental Settings}\label{subsec:a_4_expertimental_settings}
We train the LLaMA-3-8B and Mistral-7B supervised fine-tuned models on the UltraFeedback dataset using DPO and SimPO methods~\cite{touvron2023llama}.
For LLaMA-3-8B, DPO training is conducted with a beta value of 0.01 and learning rate of 5e-7, while SimPO training uses beta values of 2.0, gamma-beta ratio of 0.5, and learning rate of 6e-7. 
For Mistral-7B, DPO training is performed with beta values of 0.01 and learning rate of 5e-7, whereas SimPO uses a beta of 2.0 and a gamma-beta ratio of 0.8, and learning rate of 3e-7. 
We adopted the hyperparameter configuration from \citet{meng2024simpo}.
The model was trained for 3 epochs, and the best performance observed during training was reported.
For any process requiring random selection, such as dataset sampling, we used a fixed random seed of 42 to ensure reproducibility.
All training is conducted on four RTX A6000 ada GPUs.

\begin{table}[!ht]
  \centering
  
  \begin{tabular}{l|ccc}
    \toprule
    \textbf{Model (Method)} & $\beta$ & $\gamma/\beta$ & $lr$ \\ 
    \midrule
    LLaMA-3-8B (DPO) & 0.01 & - & 5e-7 \\
    Mistral-7B (DPO) & 0.01 & - & 5e-7 \\
    LLaMA-3-8B (SimPO) & 2.0 & 0.5 & 6e-7 \\
    Mistral-7B (SimPO) & 2.0 & 0.8 & 3e-7 \\
    \bottomrule
  \end{tabular}
  \caption{Hyperparameter used in DPO and SimPO training.}
  \label{tab:hyper_table}
\end{table}

\section{Preference Alignment by SimPO}\label{sec:b_preference_alignment_by_simpo}
\noindent
\textbf{SimPO Gradient Analysis.}
The gradient of the SimPO loss function \( \mathcal{L}_{\text{SimPO}} \) can be expressed as follows~\cite{meng2024simpo}.

{\small
\begin{align}
\nabla_\theta \mathcal{L}_{\text{SimPO}}(\pi_\theta) &= \notag -\beta \, \mathbb{E}_{(x, y_w, y_l) \sim \mathcal{D}} \left[
\sigma(-\rho_\theta) \cdot d_\theta
\right]
\end{align}
}

\noindent where
{\small

\[
\rho_\theta = 
\left(
\frac{\beta}{|y_w|} \log \pi_\theta(y_w \mid x)
- \frac{\beta}{|y_l|} \log \pi_\theta(y_l \mid x)
- \gamma
\right)
\]
\[
\quad
d_\theta = \frac{1}{|y_w|} \nabla_\theta \log \pi_\theta(y_w \mid x)
- \frac{1}{|y_l|} \nabla_\theta \log \pi_\theta(y_l \mid x)
\]
}

\begin{itemize}
    \item In \(\sigma(-\rho_\theta)\), \( \rho_\theta \) denotes the difference in log probabilities between the chosen and rejected responses. 
    When \( \rho_\theta \) is small—i.e., when the model has similar confidence in both choices—\(\sigma(-\rho_\theta)\) takes a large value. 
    Conversely, when \( \rho_\theta \) is large, \(\sigma(-\rho_\theta)\) approaches zero.
    
    \item This indicates that samples where the model is less confident in preferring the chosen response over the rejected one contribute more to the loss and result in larger gradient updates. 
    Empirically, it has been observed that for near-deterministic preference pairs (e.g., with probabilities close to 0 or 1), the gradient contribution becomes negligible, potentially limiting SimPO’s effectiveness in such scenarios~\cite{azar2024general}.
    
    \item \(d_\theta\) corresponds to the margin of the policy’s log probability gradient.
    It is well known that lower probability value result in larger gradient magnitudes~\cite{ko2024distillm}. 
    Specifically, for small values of \(\log \pi_\theta(y\mid x) \), the gradient \( \nabla_\theta \log \pi_\theta(y\mid x) \) becomes large.
    As a result, in the SimPO objective, samples with very low policy probabilities \(\log \pi_\theta(y\mid x) \) can induce excessively large gradients. 
    This may lead to unstable training behavior due to over-amplified updates in response to small preference margins~\cite{houliston2024uncertainty}.
\end{itemize}

\noindent
\textbf{Analysis on experiment results.}
We analyze the impact of these three data regions on the SimPO loss gradient in Figure \ref{fig:B_1_simpo_gradient_analysis}. 
In the \textbf{\textit{HighAvg.}} region, due to low variance and high average preference scores, both the win and loss samples tend to receive high log probability estimates. 
In contrast, the \textbf{\textit{LowAvg.}} region exhibits low variance and low average scores, resulting in low log probabilities for both win and loss samples. 
As shown in Figures \ref{fig:B_1_simpo_gradient_analysis}(a) and \ref{fig:B_1_simpo_gradient_analysis}(b), both the chosen and rejected responses exhibit higher initial log probabilities in the \textbf{\textit{HighAvg.}} region. 
This indicates that the generation probabilities for these instances are relatively higher compared to those in the \textbf{\textit{LowAvg.}} region. 
Consequently, the gradient norm behavior observed in Figure \ref{fig:B_1_simpo_gradient_analysis}(c) differs accordingly: in the \textbf{\textit{HighAvg.}} region, the gradients at the early stages of training are smaller and more stable, whereas in the \textbf{\textit{LowAvg.}} region, the gradients tend to be larger and more volatile during early training.

\section{VLFeedback Experiment}\label{sec:c_vlfeedback_experiment}
\begin{figure}[t]
    \centering
    \includegraphics[width=\columnwidth]{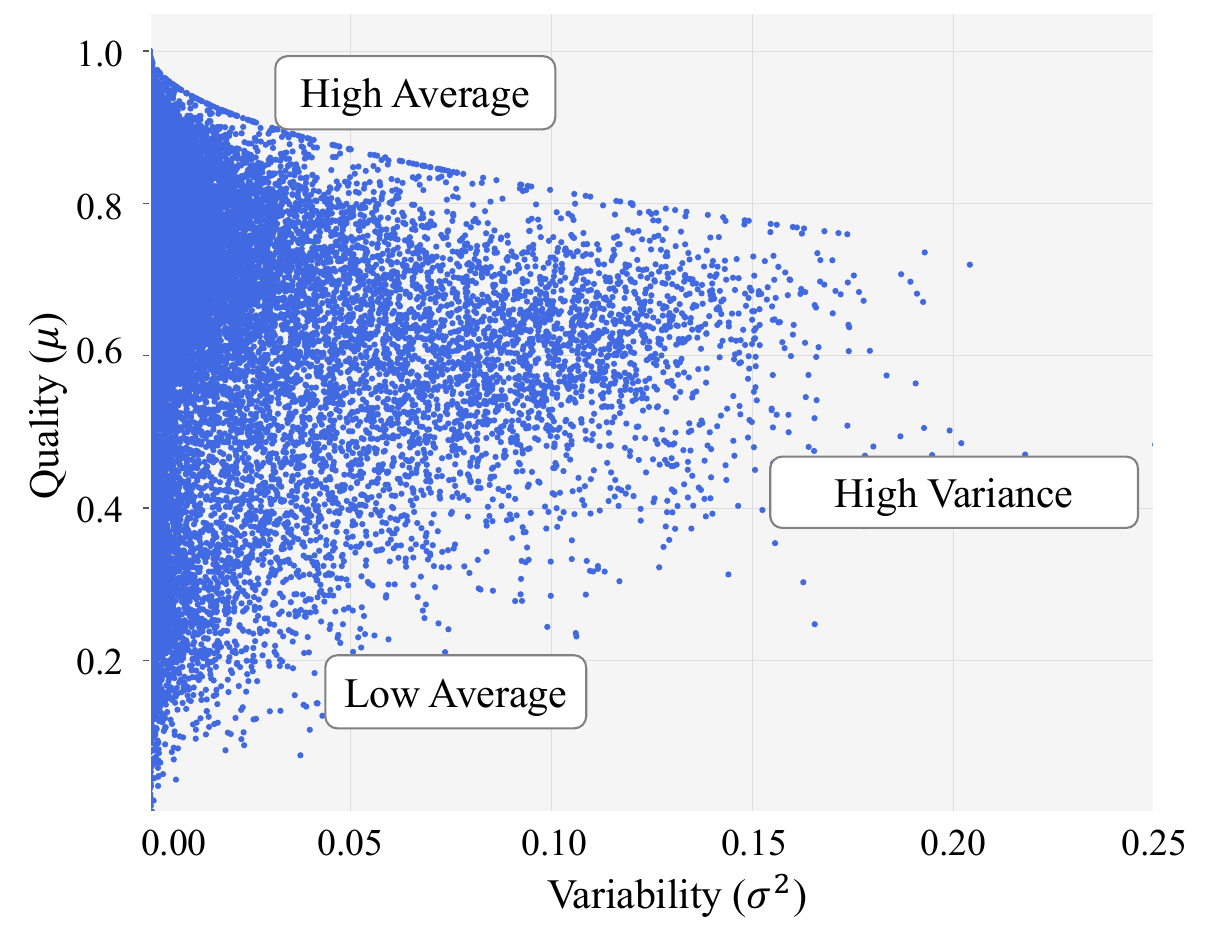}
    \caption{Alignment Data Map for the VLFeedback dataset~\cite{li2024vlfeedback}.}
    \label{fig:2_scatter_plot_vl}
\end{figure}

To extend the evaluation conducted on UltraFeedback, we perform additional data selection experiments on the multimodal visual question answering dataset \textbf{VLFeedback}~\cite{li2024vlfeedback}, in order to assess the effectiveness of our method in a multimodal setting.
For the VLFeedback experiments, we use Qwen2-VL-2B-Instruct~\cite{wang2024qwen2} as the backbone model for multimodal large language model (MLLM) alignment training.
We first construct an Alignment Data Map by applying the reference-based score, where reference answers are generated using \texttt{GPT-4V} and alignment scores are computed as cosine similarity between candidate responses and reference answers in the embedding space.
Figure~\ref{fig:2_scatter_plot_vl} illustrates the Alignment Data Map constructed on the VLFeedback dataset.

\begin{table}[!ht]
\centering
\begin{tabular}{c|l|c|c}
\toprule
\textbf{\% Train} & \textbf{Train set} & \textbf{MMBench} & \textbf{MMMU} \\
\midrule
0   & Zeroshot   & 0.720 & 41.1 \\
\midrule
\multirow{4}{*}{33} 
    & Random     & 0.719 & 42.0 \\
    & LowAvg.    & \underline{0.723} & 42.2 \\
    & HighVar.   & 0.720 & 42.1 \\
    & HighAvg.   & \textbf{0.727} & \underline{42.6} \\
\midrule
100 & Full       & 0.716 & \textbf{42.8} \\
\bottomrule
\end{tabular}
\caption{Evaluation results of our method on the VLFeedback dataset. The Qwen2-VL-2B model was trained using DPO and evaluated on MMBench~\cite{liu2024mmbench} and MMMU~\cite{yue2024mmmu}.}
\label{tab:vlfeedback_qwen2}
\end{table}

\subsection{Experimental Settings.}\label{sec:c_1_vlfeedback_experiment_settings}
We train the Qwen2-VL-2B-Instruct model on the VLFeedback dataset using the DPO method with LoRA fine-tuning. 
Training is performed with a beta value of 0.1, LoRA rank of 8, and a learning rate of 1e-6. The model is trained for 3 epochs, following the DPO implementation with sigmoid preference loss. Also, we follow the hyperparameter configuration provided in \citet{zheng2024llamafactory} VLLM settings. A fixed random seed of 42 is used for dataset sampling to ensure reproducibility. Experiments are conducted using four RTX A6000 Ada GPUs.

To assess MLLM alignment, we employ the following two alignment benchmarks:
\begin{itemize}
    \item MMBench~\cite{liu2024mmbench} is a fine-grained benchmark with 3,000 multiple-choice questions across 20 ability types. It evaluates multimodal models using an LLM-based judge to assign answer labels, supporting consistent and scalable assessment.

    \item Massive Multi-discipline Multimodal Understanding and Reasoning (MMMU)~\cite{yue2024mmmu} is designed to evaluate MLLMs on large-scale tasks that require university-level expertise and deep reasoning.
\end{itemize}

\begin{figure*}[t]
    \centering
    \includegraphics[width=\textwidth]{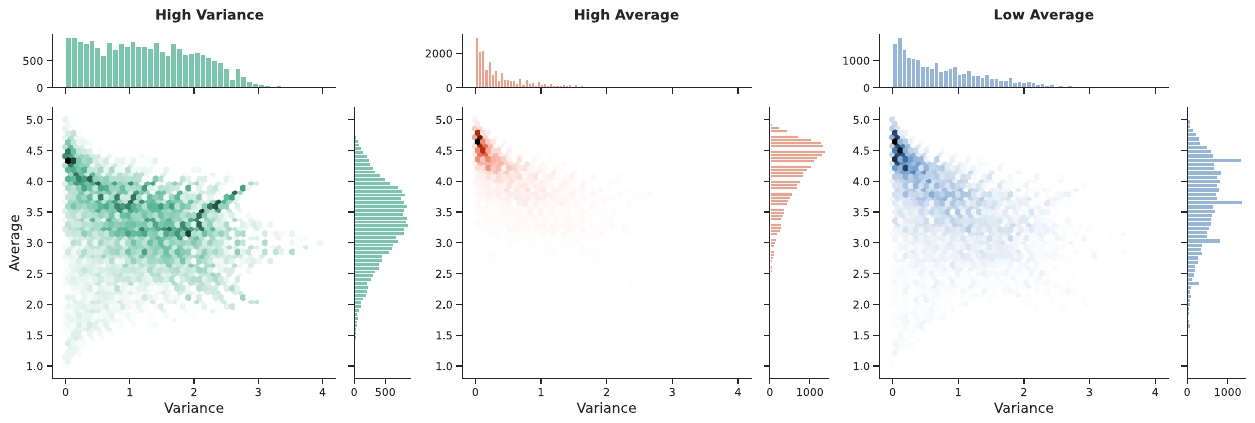}
    \caption{Hexbin jointplots for High Variance, High Average and Low Average.}
    \label{fig:C_hexbinplot}
\end{figure*}



\subsection{Experimental Results}\label{sec:c_2_vlfeedback_experiment_results}
Table~\ref{tab:vlfeedback_qwen2} presents the experimental results on the VLFeedback dataset.
For the Qwen-2-VL model, the model trained on the \textit{HighAvg.} region demonstrates strong performance, achieving a score comparable to that of the fully trained model in MMMU. \textit{HighAvg.} even surpasses the performance of fully trained models while outperforming other regions trained with 33\% of the data in MMBench. 
These findings emphasize the significant impact of dataset partitioning on MLLM training outcomes and confirm that among the three data regions, the \textit{partially aligned} region is the most effective for improving model performance.
We demonstrate that our method is robust and applicable across different data modalities and task types.


\section{Correlation Hexbin Plot}\label{sec:f_correlation_hexbin_plot}
We construct the data map using the UltraFeedback dataset, dividing it into three subsets—High Variance, High Average, and Low Average—based on the proposed criteria. For each subset, we compute the variance and average from the feedback scores $\mathcal{F}$ and visualize the resulting structure.
Figure \ref{fig:C_hexbinplot} illustrates the distribution of each subset. The High Variance group is characterized by a larger number of samples with relatively high variance when calculated using the actual preference scores. In the High Average group, most data points exhibit an average score of approximately 4.5 or higher, with variance values concentrated near zero. In contrast, the Low Average group displays a distribution skewed toward lower average scores, and the overall mean is noticeably lower than that of the High Average group.
This visualization demonstrates that the proposed data partitioning criteria effectively capture the statistical characteristics of the actual preference scores.

\begin{table*}[t]
\centering
\begin{tabular}{p{16cm}}
\toprule
\textbf{Question:} Edit the following paragraph to correct any grammatical errors: \textit{She didn't remembre where is her purse, so I thinks its in the car but he's say it's on kitchen table but he are not sure, and then they asked me to looking for it, she's say, "Can you?", and I responds with, "Maybe, but ain't no sure,” and he not heard me, and, "What?", he asks, "Did you found it?"} \\
\midrule
\end{tabular}

\vspace{0.5em}

\begin{tabular}{p{1.8cm}|p{11.5cm}|c}
\toprule
\textbf{Region} & \textbf{Model Response} & \textbf{Score} \\
\midrule
\textbf{HighAvg.} &
She couldn't remember where her purse was, so I suggested that it might be in the car. However, he said it was on the kitchen table but wasn't sure. They asked me to help find it, and she said, "Can you?" I replied, "I'm not sure, but I'll try," and he didn't hear me. He asked, "Did you find it?" "No, I'm sorry," I answered. & 9 \\
\midrule
\textbf{HighVar.} &
It is not clear from the sentence "she's say, `Can you?' and I responds with, 'Maybe, but ain't no sure,'" what the speaker is responding to. It could be interpreted as a response to the question "Can you?" or to "Did you found it?" It is also unclear if the speaker is responding to a question being asked by another person or is responding to their own thoughts. To correct these grammatical errors, the sentence could be rewritten as: "She asked me, 'Can you find my purse?' and I replied, `I'm not sure, but I'll try my best.' & 4 \\
\midrule
\textbf{LowAvg.} &
It is not clear from the sentence "she's say, `Can you?' and I responds with, `Maybe, but ain't no sure,'" what the speaker is responding to. It could be interpreted as a response to the question "Can you?" or to "Did you found it?" It would be helpful to clarify which question the speaker is responding to. Additionally, the sentence "he not heard me" should be changed to "he didn't hear me" for proper grammar. Finally, the sentence "and, `What?’, is not necessary as it is not a complete sentence. & 4 \\
\bottomrule
\end{tabular}
\caption{Case study from MT-Bench evaluating LLaMA-3-8B model fine-tuned via SimPO. Responses from different data regions (High Average, High Variance, Low Average) are compared in terms of grammatical correction quality.}

\label{tab:A_D_case_study_grammar}
\end{table*}

\section{Case Study}\label{sec:e_case_study}
To verify whether models trained on different data regions exhibit distinct behaviors, we conduct a case study comparing their actual responses.
Table \ref{tab:A_D_case_study_grammar} presents a case study comparing model responses across three data regions: \textbf{\textit{High Average}} \textit{(HighAvg.)}, \textbf{\textit{High Variance}} \textit{(HighVar.)}, and \textbf{\textit{Low Average}} \textit{(LowAvg.)}.
Given a prompt requiring grammatical correction, the HighAvg response provides a clear and coherent revision with a high score of 9.
In contrast, both \textit{HighVar.} and \textit{LowAvg.} responses receive lower scores of 4 due to ambiguity, insufficient grammatical corrections, and failure to follow the instruction precisely. 
These responses are also less aligned with human preferences in terms of clarity and task relevance.
Specifically, these responses fail to clarify the speaker's intent and do not fully resolve key errors in the original text.
These cases illustrate that data from the \textit{HighAvg.} region better supports instruction-following and clarity, while low-quality or inconsistent data leads to suboptimal model behavior.

\section{Validation Against Human Preferences}\label{sec:F_validation}
To validate whether alignment scores are consistent with human-labeled preferences, we compare them against the human annotations in the Preference-Dissection dataset.
Human annotations are provided in a pairwise format, where annotators select the preferred response between two candidates for each prompt.
We define agreement as the case where the response preferred by humans receives a higher alignment score.
Under this definition, we observe an agreement rate of approximately 68\% over 5.2k samples.

To further assess the reliability of alignment scores in cases of disagreement, we construct two training datasets: (1) one using human preference annotations and (2) one using alignment-score-based preferences.
We then train separate models on each dataset and compare their performance.

\begin{table}[!t]
\centering
\resizebox{\columnwidth}{!}{
\footnotesize
\begin{tabular}{ll|cc|cc}
\toprule
\multicolumn{2}{c|}{\textbf{Method}} 
& \multicolumn{2}{c|}{\textbf{MT-Bench}} 
& \multicolumn{2}{c}{\textbf{Evol-Instruct}} \\
\cmidrule(lr){3-4} \cmidrule(lr){5-6}
& 
& \textbf{WR} & \textbf{Single} 
& \textbf{WR} & \textbf{Single} \\
\midrule
\multicolumn{2}{c|}{Zero-Shot} 
& 0.278 & 4.61 
& 0.234 & 5.80 \\
\midrule
Right (Acc = 1) 
& PD = AS
& 0.272 & 4.53 
& 0.250 & 5.98 \\
\midrule
\multirow{2}{*}{Wrong (Acc = 0)} 
& PD 
& 0.281 & 4.50 
& 0.236 & 5.86 \\
& AS
& \textbf{0.319} & \textbf{4.56} 
& \textbf{0.257} & \textbf{5.96} \\
\bottomrule
\end{tabular}
}
\caption{\label{tab:pref_dissection_result}
Evaluation results on MT-Bench and Evol-Instruct. WR denotes win rate and Single denotes the average score. Bold indicates the best performance within the Wrong (Acc = 0) subset.}
\vspace*{-1em}
\end{table}

Consistent with the correlation trends observed in the main paper, alignment-score-based labels yield better performance than Preference-Dissection labels on the subset of samples where disagreements occur.
Overall, while alignment scores do not perfectly match human preferences, the level of agreement is substantial, and the additional experiments suggest that alignment-score-based labels remain a reliable signal for preference learning despite some mismatches.
These findings further support the robustness of our correlation analysis and its applicability across different datasets.

\end{document}